\documentclass{IEEEtran}
\pdfoutput=1
\usepackage[cmex10]{amsmath}
\usepackage{amssymb}
\usepackage[pdftex]{graphicx}
\usepackage{times}
\usepackage{hyperref}

\newcommand{\IMPROVINGNEIGHBOR}{\textsc{Improving-Neighbor}}
\newcommand{\XX}{\mathcal X}
\newcommand{\Nl}{N_\text{layers}}
\newcommand*\de{\mathop{}\!\mathrm{d}}
\newcommand{\wmax}{w_\text{max}}

\begin{document}


\title{A Telescopic Binary Learning Machine for Training Neural Networks}

\author{\IEEEauthorblockN{Mauro Brunato}
and
\IEEEauthorblockN{Roberto Battiti}\\
\IEEEauthorblockA{DISI --- University of Trento, \\
Via Sommarive 9, I-38123 Trento, Italy \\
Email: \{brunato,battiti\}@disi.unitn.it}
}

\maketitle


\begin{abstract}

This paper proposes a new
algorithm based on multi-scale stochastic local search with binary representation
for training neural networks.

In particular, we study the effects of neighborhood evaluation strategies, the effect of the number of bits per
weight and that of the maximum weight range used for mapping binary strings
to real values. Following this preliminary investigation, we propose a telescopic
multi-scale version of local search where the number of bits is increased
in an adaptive manner, leading to a faster search and to local minima
of better quality. An analysis related to adapting the number of bits
in a dynamic way is also presented.
The control on the number of bits,
which happens in a natural manner in the proposed method, is effective to increase the generalization
performance.

Benchmark tasks include a highly non-linear artificial problem, a control
problem requiring either feed-forward or recurrent architectures for feedback control,
and challenging real-world tasks in different application domains.
The results demonstrate
the effectiveness of the proposed method.

\end{abstract}



\section{Introduction}

Machine learning and optimization share a long common history:
machine learning implies optimizing the performance
of the trained system measured on the examples, with possible early stopping or additional
regularizing terms
in the function to avoid overtraining and increase generalization.

Starting from  back-propagation \cite{werbos1974beyond,Rumelhart86},
most techniques for training neural networks use continuous optimization schemes based
on partial derivatives, like gradient descent and variations thereof, while
Support Vector Machines (SVMs) use quadratic optimization \cite{svm}.
Recent proposals to simplify the optimization task
by creating a randomized first layer and limiting optimization to the last-layer weights
are extreme learning \cite{huang2006extreme} and reservoir computing \cite{lukovsevivcius2009reservoir}.

Different methods consider Combinatorial Optimization (CO) techniques without
derivatives, like Simulated Annealing and Genetic Algorithms, see \cite{sexton1999optimization}
for a comparative analysis. In most CO techniques, weights are considered as binary
strings and the operators acting during optimization change bits, either individually (like with
mutation in GA) or more globally (like with cross-over operators in GA).
Methods for the optimization of functions of real variables with no
derivatives are an additional option, for example direct search \cite{hooke1961direct}
or versions of Simulated Annealing for functions of continuous variables \cite{corana87}.
Intelligent schemes based on adaptive diversification strategies by prohibiting
selected moves in the neighborhood are considered in \cite{BaTe95b}.

A strong motivation for considering techniques with no derivatives is when
derivatives are not always available, for example in threshold networks \cite{barlett1992using}, because
their transfer functions are
discontinuous.
Preliminary results for threshold networks are presented in \cite{bb2015ijcnn}.


This paper aims at considering neural networks with smooth input-output transfer functions
and proposes a new method, called Telescopic Binary Learning Machine (BLM for short), that combines
stochastic local search optimization with the discrete representation of network parameters (weights).
The technique is suitable both for feed-forward and recurrent networks, and can also be applied to the control
of dynamic systems with feedback.

The paper is organized as follows: in Sec.~\ref{sec:LS} we introduce the SLS techniques that will be used
for the optimization of network parameters; in Sec.~\ref{sec:BLM} we describe the specific building blocks
and algorithmic choices that enable the BLM algorithm; in Sec.~\ref{sec:two-spirals} we use a highly non-linear
benchmark task to develop insight and drive the choice of specific algorithms and parameters; in Sec.~\ref{sec:pendulum}
we apply the network to a well-known control problem and test the suitability of BLM for training recurrent networks;
in Sec.~\ref{sec:feedforward} we test the technique on a set of classic real-world benchmark problems and compare it with
state-of-the-art derivative-based and derivative-free methods. Finally, in Sec.~\ref{sec:conclusions}
we draw some conclusions and indicate possible further developments.









\section{Local search on highly structured fitness surfaces}
\label{sec:LS}

Local search is a basic building block
of most heuristic search methods for combinatorial optimization.
It consists of starting from an initial tentative solution and trying to
improve it through repeated small changes. At each repetition the current is
\emph{perturbed} and the function to be optimized is tested.
The change is kept if the new
solution is better, otherwise another change is tried. The function $f(X)$ to be optimized is called
\emph{fitness} function, \emph{goodness} function, or
\emph{objective} function.

In a way, gradient-based techniques like back-propagation (BP) also consider the effect of small
local changes. In BP the effect of the tentative change is approximated by the local linear
model: the scalar product between gradient and displacement.
When no analytic derivatives are available, the derivatives can
be approximated by finite differences. One is therefore dealing with
techniques based on gradual changes evaluated through their effect on the
error function, but considering individual bits leads to a radical simplification,
and to a natural multi-scale structure because of the binary representation
(change of the $i$-th bit leads to a change proportional to~$2^i$).

Let's define the notation.
$\XX$ is the search space,  $f(X)$  the function to be optimized,  $X^{(t)}$ is the current solution at
iteration (``time'') $t$. $N(X^{(t)})$ is the neighborhood of point $X^{(t)}$, obtained by applying a
set of basic moves $\mu_0, \dots, \mu_M$ to the current configuration:
$$
	N(X^{(t)}) = \{ X \in \XX \quad : \quad X = \mu_i (X^{(t)}), i=0,\dots,M \} .
$$
If the search space is given by binary strings with a given length $L$, then $\XX = \{0,1\}^L$,
the moves can be those changing (or complementing or \emph{flipping}) the individual bits, and
therefore $M$ is equal to the string length $L$.

\emph{Local search} starts from an admissible configuration $ X^{(0)}$ and builds a
\emph{trajectory} $X^{(0)},\dots,X^{(t+1)}$. The successor of the
current point is a point in the neighborhood with a lower value of the function $f$ to be minimized:
\begin{eqnarray*}
	Y & \leftarrow & \IMPROVINGNEIGHBOR \bigl(N(X^{(t)})\bigr) \\
	X^{(t+1)} & = & \begin{cases}
		Y  & \text{if $f(Y) < f(X^{(t)})$} \\
		X^{(t)} & \text{otherwise (search stops)}.
	\end{cases}
\end{eqnarray*}
Function \IMPROVINGNEIGHBOR\ returns an improving element in the neighborhood. In a simple case this is
the element with the lowest $f$ value, but other possibilities exist, for
example the \emph{first improving} neighbor encountered.
If no neighbor has a better $f$ value, i.e., if the configuration is a \emph{local minimizer},
the search stops.

Local search is surprisingly effective because most combinatorial optimization problems have a very
\emph{rich internal structure} relating the configuration $X$ and the $f$ value.
If one starts at a good solution, solutions of similar quality can, on the
average, be found more \emph{in its neighborhood} than by sampling a completely unrelated random
point. A \emph{neighborhood} is suitable for local search if it reflects the problem structure.
Stochastic elements can be immediately added to local search to make it more
robust \cite{Hoos2005}, for example a random sampling of the neighborhood can be adopted
instead of a trivial exhaustive consideration.

\begin{figure}[tb]
	\centering
	\includegraphics[width=\columnwidth]{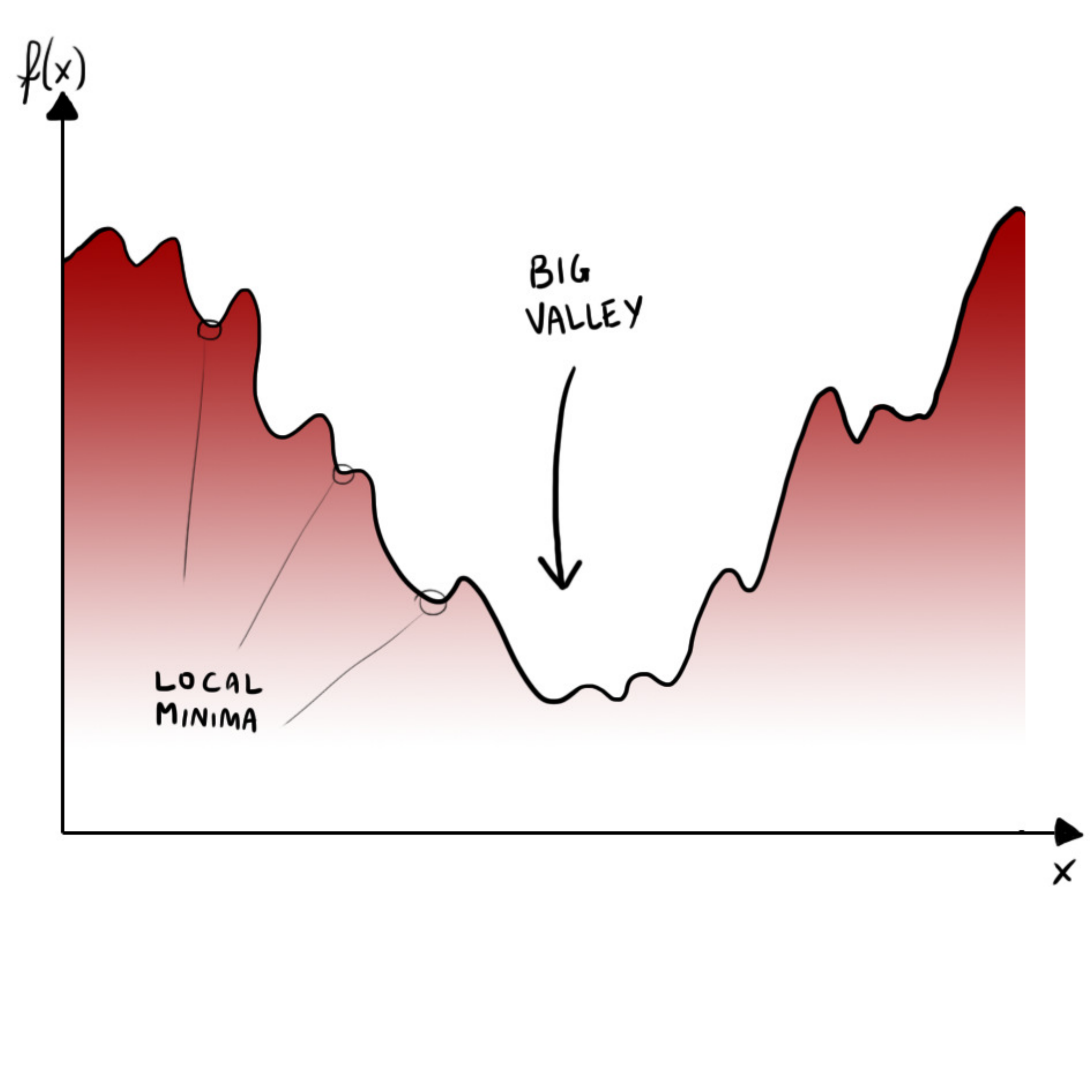}
	\vskip-2cm
	\caption{The \emph{massif central} hypothesis: local minima tend to occur in clusters (from~\cite{lionway}).}
	\label{fig:massif}
\end{figure}

In many relevant problems local minima tend to be \emph{clustered},
furthermore good local minima tend to be closer to other good minima.
Let us define as \emph{attraction basin}
associated with a local optimum the set of points $X$ which are mapped to the given local optimum by the
local search trajectory.
One may think about a smooth $f$ surface in a continuous environment, with basins
of attraction which tend to have a nested, ``fractal'' structure, like in Fig.  \ref{fig:massif}.
This \emph{multi-scale} structure, where smaller
valleys are nested within larger ones, is the basic motivation for more complex
methods like Variable Neighborhood
Search, Iterated Local Search (ILS), see for example~\cite{BBM2008thebook}.
This structural property is also called \emph{Big Valley property} (or \emph{massif central}).

In Simulated Annealing, a tentative move is generated and accepted with a probability
depending on the difference in value $\Delta f$ caused by the move:
$$
	\Pr(\text{acceptance}) \approx e^{-\frac{\Delta f}T}.
$$
Studies and experiments of Simulated Annealing \cite{KGV83,laa87} motivate cooling schedules (gradual reduction
of the $T$ parameter controlling move acceptance) so that large temperature at the beginning
causes a sampling of states regulated by the overall large-scale structure of the problem,
while lower temperature at the end makes the search more sensitive to fine-scale details.
The role of high temperatures is to ``iron out''
the fine details of the fitness surface at the beginning, to avoid the search being
trapped in a poor local minimum.

Multi-scale methods which jointly solve related versions of the same problems
at different scales and have been proposed in different areas, e.g.
image analysis \cite{starck1998image}, modeling and computation \cite{weinan2003multiscale},
neural networks for optimization \cite{mjolsness1991multiscale}. The application in \cite{mjolsness1991multiscale}
is for multi-scale optimization (inexact graph matching). A cubic
neural network energy function is considered, a smaller coarser version of the problem
is derived, and one alternates between relaxation steps for the fine-scale
and coarse-scale problems.

In this paper, we consider two simple options: no diversification (where the search stops upon reaching a local optimum),
and \emph{repeated local search}, where the search restarts from a random configuration whenever a local optimum is attained.

In the following, the term \emph{run} will refer to a sequence of moves that ends to a local minimum of the search space, while
a \emph{search} can refer to a sequence of runs and restarts.

In the following we illustrate the main building blocks leading to a much faster and more
effective multi-scale implementation of local search for neural networks.


\section{The Binary Learning Machine (BLM) Algorithm}
\label{sec:BLM}

A brute-force implementation of local search consists of evaluating
all possible changes of the individual bits representing the weights.
For each bit change all training examples are evaluated to calculate
the difference in the training error.
This implementation is out of question for networks
with more than a few neurons and weights, leading to enormous CPU times.
This can explain why basic local search has
not been considered as a competitive approach for training up to now.

In the following sections we design an algorithm, called 
Telescopic Binary Learning
Machine (\emph{BLM} for short), which uses a smart realization of local stochastic search (SLS),
  coupled with an efficient network representation. The term Binary Learning Machine  underlines the fact
that it is based on the binary representation of weights and changes acting on the individual bits.

Telescopic BLM is based on:

\begin{itemize}
 \item Gray coding to map from binary strings to weights
 \item Incremental neighborhood evaluation (also called delta evaluation)
 \item Smart sampling of the neighborhood (stochastic ``first improving'' and ``best improving'')
 \item Multi-scale (telescopic) search, gradually increasing the number of bits
in a dynamic and adaptive manner
\end{itemize}

In this Section we illustrate the many choices to obtain a representation of the neural network
that can be efficiently optimized by a SLS algorithm.

\subsection{Weight representation}

The appropriate choice of representation is critical to ensure that changes of individual
bits lead to an effective sampling of the neighborhood.

\subsubsection{Number of bits and discretization}
We represent weights and biases as integral multiples of a discretization value $\epsilon>0$, with an $n$-bit
two's-complement integer as multiplier. Options $n=2,4,8,12,16,24$ have been investigated.
In all tests a maximum representable weight value $\wmax$ is given
(this limitation in weight size acts as an intrinsic regularization mechanism to prevent overtraining), and
$\epsilon$ is set so that the range of representable weights
includes the interval $[-\wmax,\wmax]$.
As the two's-complement representation is asymmetric, with $n$ bits actually representing the set
$\{-2^{n-1},\dots,2^{n-1}-1\}$, we set
\begin{equation}
	\epsilon = \frac\wmax{2^{n-1}-1}
	\label{eq:discretization step}
\end{equation}
so that the maximum value $\wmax$ can be represented;
the minimum representable weight will actually be $-\wmax-\epsilon$.
Since we start from $n=2$ bits, one can always represent at least one positive and two negative weight values.

\begin{figure}[tbp]
    \includegraphics[width=\hsize]{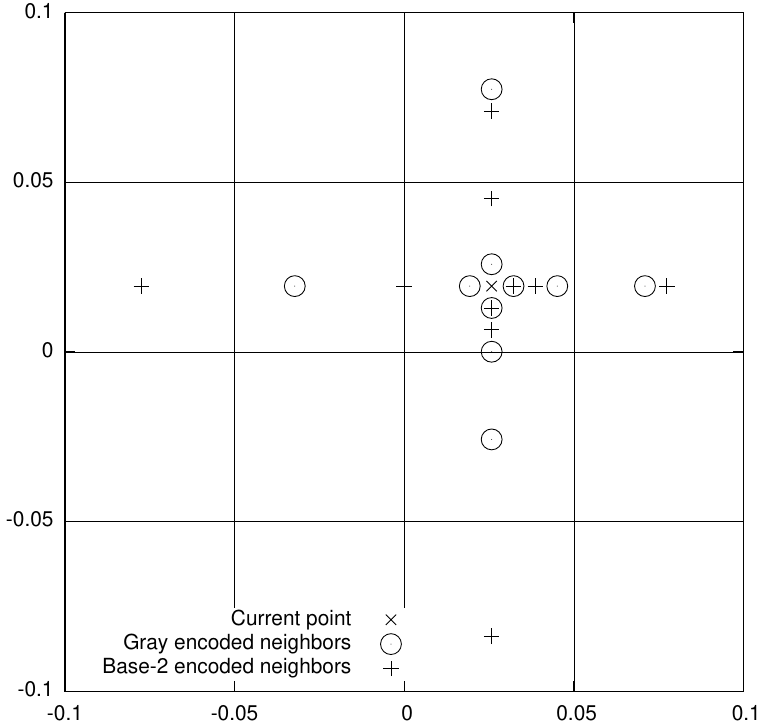}
    \caption{Neighbors of the current point with Gray coding vs. standard base-2 coding.}
    \label{fig:Gray}
\end{figure}
\subsubsection{Gray coding}
In order to improve the correspondence between the bit-flip local moves of the BLM algorithm and the
natural topology of the network's weight space, we use Gray encoding of the $n$-bit integer multiplier.
Gray encoding has the property that, starting from code $n$, the nearby
codes for $n - 1$ and $n + 1$ are obtained by changing a single bit (i.e., they have a
Hamming distance of one with respect to the code of $n$).
This choice allows the generic value $h\epsilon$, $h$ being an $n$-bit integer, to reach its neighboring values $(h+1)\epsilon$
and $(h-1)\epsilon$ by a single move, while in the conventional binary coding one of the two neighbors could be
removed by as much as $n$ moves, namely when moving from $0$ to $-1$ (represented as $00\cdots0$ and $11\cdots1$ in
two's complement standard binary).\\
Given a weight $w$, by changing one of the $n$ bits in the
encoding (and by repeating the operation for all possible bits),
one obtains $n$ weights in the neighborhood. As shown in Fig.~\ref{fig:Gray}, for the cited
property of the Gray code, the neighborhood always contains the
nearest weights on the discretized grid, plus a cloud of points
at growing distances in weight space.

\subsection{Memory-based incremental neighborhood evaluation}

Let $\Nl$ be the number of network layers (hidden and outputs); the input layer is layer~0, while layer~$\Nl$ is the output.
For $l=0,\dots,\Nl$, let $n_l$ be the number of neurons in layer $l$. For $j=1,\dots,n_l$, let $o^l_j$ be the output of neuron $j$ in layer $l$. Thus, $o^0_j$
is the value of the $j$-th input. Let $f^l(x)$ be the transfer function of neurons at layer $l$, and $w^l_{ij}$ be the weight connecting neuron $i$ at layer $l-1$
to neuron $j$ at layer $l$. Starting from the input values, the feed-forward computation consists of iterating the following operation for $l=1,\dots,\Nl$:
\begin{equation}
    \label{eq:feedforward}
    o^l_j=f^l(s^l_j),\qquad\text{where}\qquad s^l_j=b^l_j+\sum_{i=1}^{n_{l-1}}w^l_{ij}o^{l-1}_i.
\end{equation}
The simple bit-flip neighborhood scheme chosen for this investigation requires a large number of steps, each consisting of
the modification of a single weight. A very significant speedup can be obtained by storing the $s^l_j$ values for all
neurons and all samples. In order to obtain the network output when weight $w^L_{IJ}$ is replaced by a new value $W$,
the following algorithm provides a faster incremental evaluation:
\begin{enumerate}
    \item
        Recompute the output of neuron $J$ in layer $L$:
        $$
            o'^L_J = f^L\bigl(s^L_J + (W-w^L_{IJ})o^{L-1}_I\bigr).
        $$
        If $L=\Nl$, the $J$-th network output is $o'^L_J$, the others are unchanged. Stop.
    \item
        Let the output variation be
        $$
            \Delta o^L_J = o'^L_J - o^L_J.
        $$
    \item
        Recompute all contributions of output $o^L_J$ to the neurons
        of the next layer, for $k=1,\dots,n_{L+1}$:
        $$
            s'^{L+1}_k = s^{L+1}_k + w^{L+1}_{Jk}\Delta o^L_J;
        $$
        $$
            o'^{L+1}_k = f^{L+1}(s'^{L+1}_k).
        $$
    \item If there are other layers, apply~(\ref{eq:feedforward}) for $l=L+2,\dots,\Nl$.
\end{enumerate}
Obvious modifications apply for bias values.

\begin{figure}[tbp]
	\includegraphics[width=\hsize]{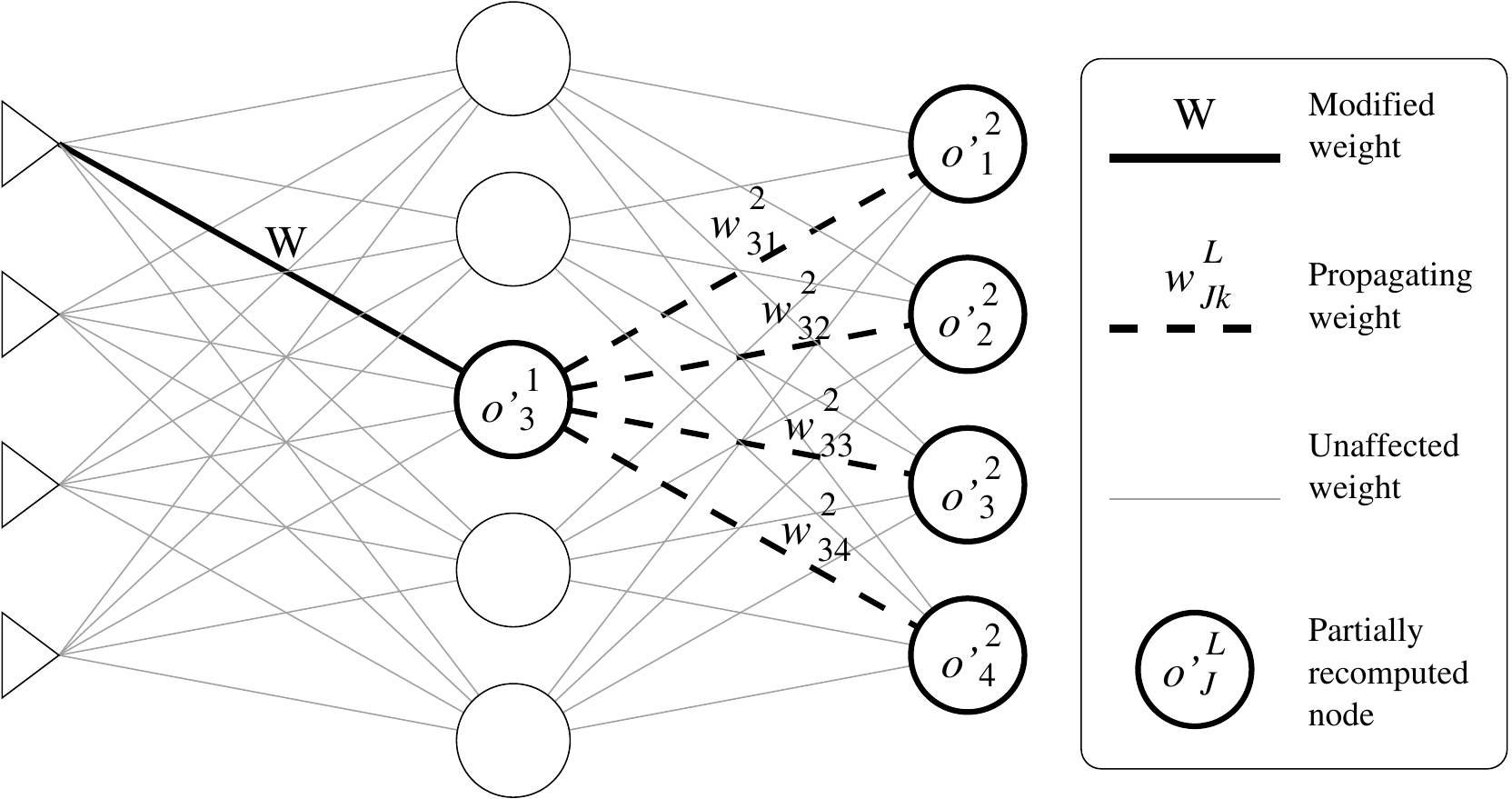}
	\caption{Affected neurons in an incremental evaluation step with a single changed weight in the input layer.}
	\label{fig:incremental evaluation}
\end{figure}
Consider the case shown in Fig.~\ref{fig:incremental evaluation},
where $\Nl=2$ (single hidden layer). Upon modification of an input weight,
the update procedure requires $O(1)$ operations per
sample to update the output of the only affected hidden neuron, and $O(n_2)$ operations
to recompute the contributions of that hidden neuron on all outputs.
Upon modification of an output weight, the whole incremental procedure requires
$O(1)$ calculations for recomputing the affected output neuron.
On average, since the network contains $O(n_0n_1)$
input weights and $O(n_1n_2)$ output weights, an incremental computation requires $O\bigl(n_0n_2/(n_0+n_2)\bigr)$ operations.

For comparison, the number of operations required by the complete feed-forward procedure is roughly proportional to the number of weights in the network, i.e., $O\bigl(n_1(n_0+n_2)\bigr)$.
An incremental search step on the largest benchmark considered is faster by about two orders of magnitude with respect
to the complete evaluation.

\subsection{Neighborhood exploration}
An important parameter for the implementation of a local search heuristic is the fraction of neighborhood
that the algorithm explores before moving, and two extreme cases are:
\begin{itemize}
    \item \emph{Best move} --- all feasible moves are checked, and the one leading to the best improvement in the
    target function is selected. Ties may be broken randomly or by considering secondary objectives.
    \item \emph{First improving move} --- feasible moves are explored in random order, and the first
    move that leads to an improvement of the target objective is selected.
\end{itemize}
Preliminary tests of the BLM algorithm
suggest that the Best move strategy causes a serious slowdown of each iteration and, while having a steeper initial
improvement, it can lead to more over-training. Therefore, in the remainder of our work we shall only consider
the First-move strategy.

Since the search starts from a random configuration, at the beginning most moves are improving, and the algorithm only needs to scan a small
subset before finding one. As the search proceeds, the average number of neighbors that must be evaluated increases.
The fraction, however, does not become very large until the local minimum is close, and remains in the $10\%$ range for most of the run.

The combination of incremental evaluation and first-improving strategy lead to an approximate speedup in the order of 100~times with respect to the baseline algorithm for the benchmark problems considered.

\subsection{Multi-scale network representation (Telescopic BLM)}
Given the potentially large number of weights in a fully connected feed-forward neural network, its discrete $n$-bit representation
can be large, leading to slow improvements. The \emph{Telescopic} variant of BLM starts with a coarser representation
where only the most significant $n'$ bits per weight (e.g., $n'=2$) are allowed to change. This allows for a quick, large-scale
exploration of the search space. Periodically, the number $n'$ of flippable bits is increased, until the full-fledged $n$-bit
search takes place.

As the search proceeds, therefore, the neighborhood structure of the problem becomes richer and richer;
moreover, the most significant bits are always allowed to flip, therefore the search algorithm is never restrained in small
regions as the search proceeds (as is often the case with backpropagation and learning-rate-decay mechanisms).

The decision to increase the number $n'$ of unlocked bits can be linked in an adaptive manner to the current search results.
The simplest criterion is to increase $n'$ whenever a local minimum of the loss function is reached for the current neighborhood structure. In this case, an additional bit is ``unfrozen'' and available for tests and possible changes.

\subsection{Telescopic threshold adjustment}
Instead of waiting until a local minimum is found,
one can increase the number of bits in the weight representation of the BLM algorithm whenever the
number of improving neighbors falls below a certain threshold (e.g., expressed as a fraction of the
total number of possible moves in the neighborhood).

In a first-improvement scenario, we cannot directly determine the desired number. Therefore, we need to
efficiently estimate it.

Below, we proceed by first analyzing the inverse problem (a), then we use the solution to
determine a threshold (b).

\subsubsection{Computing the expected number of moves before improvement}
\label{sec:inverse}
Let $A_N$ be the set of $N=2^{n_\text{weights}n'}$ local moves, $k$ of which are improving. We want to compute the
expected number of moves to be tested before we find one of the $k$ improving ones.
Moves are randomly selected, therefore we can assume that the $k$ improving moves
are randomly distributed in $A_N$. Therefore, our problem is equivalent to the following:\\
\emph{
	Let
	$$
		A_N=\{0,\dots,N-1\}
	$$
	be the set of the first $N$ natural numbers. Given the natural number $1\le k\le N$,
	we want to determine the expected value of the minimum element in subsets of $A_N$
	of cardinality $k$:
	$$
		E_{k,N} = E(\min X: X\subseteq A_N \wedge |X|=k).
	$$
}\\
Let us define $s_{m,k,N}$ as the number of subsets of $A_N$ having cardinality $k$ and minimum $m$:
$$
	s_{m,k,N}=|\{X\subseteq A_N:|X|=k\wedge\min X=m\}|.
$$
The following inductive relationship holds:
$$
	s_{m,k,N} = \begin{cases}
		\displaystyle\binom{N-1}{k-1} & \text{if $m=0$}\\
		s_{m-1,k,N-1} & \text{if $1\le m\le N-k$}\\
		0 & \text{otherwise.}
	\end{cases}
$$
In fact, if $m=0$ then $0\in X$, and to complete such a set $X$ we have to select its remaining $k-1$ elements among the remaining
$N-1$ nonzero elements of $A_N$. Moreover, the minimum of any set $X\subseteq A_N$ containing $k$ elements
is at most $N-k$, therefore if $m>N-k$ then $s_{m,k,N}=0$. In all other cases, the number of such subsets is equivalent to the number
of subsets of cardinality $k$ of  $A_{N-1}$ having minimum $m-1$ (remove $0$ from $A_N$ and subtract $1$ from all elements).\\
Therefore, let $p_{m,k,N}$ be the probability that a subset of cardinality $k$ of $A_N$ has minimum $m$:
\begin{eqnarray*}
	p_{m,k,N} &=& \Pr(\min X=m: X\subseteq A_N \wedge |X|=k)\\ &=& {\binom Nk}^{-1}s_{m,k,N};
\end{eqnarray*}
then,
\begin{eqnarray}
	&&E_{k,N} = \sum_{m=0}^{N-k} m\cdot p_{m,k,N} = {\binom Nk}^{-1}\sum_{m=1}^{N-k}ms_{m,k,N} \nonumber\\
	&&	= {\binom Nk}^{-1}\left(\sum_{m=1}^{N-k}s_{m,k,N} + \sum_{m=1}^{N-k}(m-1)s_{m-1,k,N-1}\right) \nonumber\\
	&&	=  \overbrace{{\binom Nk}^{-1}\sum_{m=1}^{N-k}s_{m,k,N}}^{\alpha_{k,N}}
			+ \overbrace{{\binom Nk}^{-1}\binom{N-1}kE_{k,N-1}.}^{\beta_{k,N}} \label{eq:ekn}
\end{eqnarray}
Let us obtain a recurrence relation for the first term in~(\ref{eq:ekn}). Clearly, $\alpha_{k,k}=0$. Otherwise, if $N>k$:
\begin{eqnarray*}
	\alpha_{k,N} &=& {\binom Nk}^{-1}\sum_{m=1}^{N-k}s_{m-1,k,N-1}\\
		&=& {\binom Nk}^{-1}\left(\binom{N-2}{k-1}+\sum_{m=1}^{N-k-1}s_{m,k,N-1}\right)\\
		&=& {\binom Nk}^{-1}\left(\binom{N-2}{k-1}+\binom{N-1}k\alpha_{k,N-1}\right)\\
		&=& \frac{N-k}N\left(\frac k{N-1} + \alpha_{k,N-1}\right).
\end{eqnarray*}
The term $\beta_{k,N}$ in~(\ref{eq:ekn}) can be simplified as
$$
	\beta_{k,N}=\frac{N+k}N E_{k,N-1}.
$$
By putting all pieces together, we obtain the recurrent relation for $E_{k,N}$:
\begin{eqnarray}
	\alpha_{k,N} &=& \begin{cases}
		0 & \text{if $N=k$}\\
		\displaystyle\frac{N-k}N\left(\frac k{N-1} + \alpha_{k,N-1}\right) & \text{otherwise}
	\end{cases}\label{eq:alpha}\\
	E_{k,N} &=& \begin{cases}
		0 & \text{if $N=k$}\\
		\displaystyle\alpha_{k,N} + \frac{N+k}N E_{k,N-1} & \text{otherwise.}
	\end{cases}\label{eq:ekn final}
\end{eqnarray}

\subsubsection{Application to telescopic BLM}
\label{sec:threshold}
Let $c_i$ be the number of moves that have been sequentially tested at step $i$ of the local search
procedure before finding an improving one; in other words, the $c_i$'s play the role of index $m$
in the previous section. If we assume that the number of improving moves
does not change dramatically between iterations, then a moving average of the $c_i$'s is an
estimator of $E_{k_i,N}$, where $k_i$ is the (unknown) total number of improving moves at time $i$:
\begin{equation}
	E_{k_i,N} \approx \mu_i,
	\label{eq:ekn approx}
\end{equation}
where $\mu_0=0$, $\mu_i=\eta\mu_{i-1}+(1-\eta)c_i$ is an exponential moving average with decay factor $\eta$.\\
Suppose that we want to increase the number of bits in the search space whenever the number of improving moves
$k_i$ falls below a given threshold $\phi N$ (given as a fraction of the total number of moves $N$, with $0\le\phi\le1$);
since $E_{k,N}$ is a decreasing function of $k$, then the condition $k_i<\phi N$ is equivalent to
$$
	E_{k_i,N}\ge E_{\lfloor\phi N\rfloor,N};
$$
finally, by taking into account the above approximation~(\ref{eq:ekn approx}), the telescopic criterion
becomes
$$
	\mu_i\ge E_{\lfloor\phi N\rfloor,N}.
$$
In order to use this criterion, we need to recompute the threshold $E_{\lfloor\phi N\rfloor,N}$
by using~(\ref{eq:alpha}) and~(\ref{eq:ekn final}) whenever the
number $N$ of moves changes (i.e., when the number of bits $n'$ in the weights representation is increased),
and maintain a mobile average of failed attempts per local move with a suitable decay factor $\eta$.\\
Observe that the initial value for the moving average, $\mu_0=0$, corresponds to the optimistic assumption that
all moves are improving: this choice mitigates the impact of high $c_i$ values happening by chance
at the beginning of the search. The moving average is reset to zero every time the number of bits is increased.

\subsection{Weight initialization}
The single-bit-flip move structure presented above is not directly compatible with all-zero weight initialization.
As soon as the network has at least one hidden layer, in fact, at least two non-zero weights, one in the hidden
and one in the output layer, are necessary to have non-constant output.

\subsubsection{Full random initialization}
The simplest initialization strategy is to initialize all weights within their variation range, using a uniform
distribution. This is equivalent to initializing every bit of the binary representation to~$0$ or~$1$ with equal
probability.

\subsubsection{Bounded random initialization}
Starting from small random weight values is often advisable: small weights lead to better generalization, and this is particularly important at
the start of training, where weights shouldn't be strongly polarized. Therefore, setting a small initialization range $[-w_\text{init},w_{init}]$
is a valid choice. Since weights are discretized, it's important that the maximum initial weight is at least as large as the discretization step:
$w_\text{init}\ge\epsilon$. Initial weights are then rounded to the closest discretized value.

\begin{figure}[tbp]
	\includegraphics[width=\hsize]{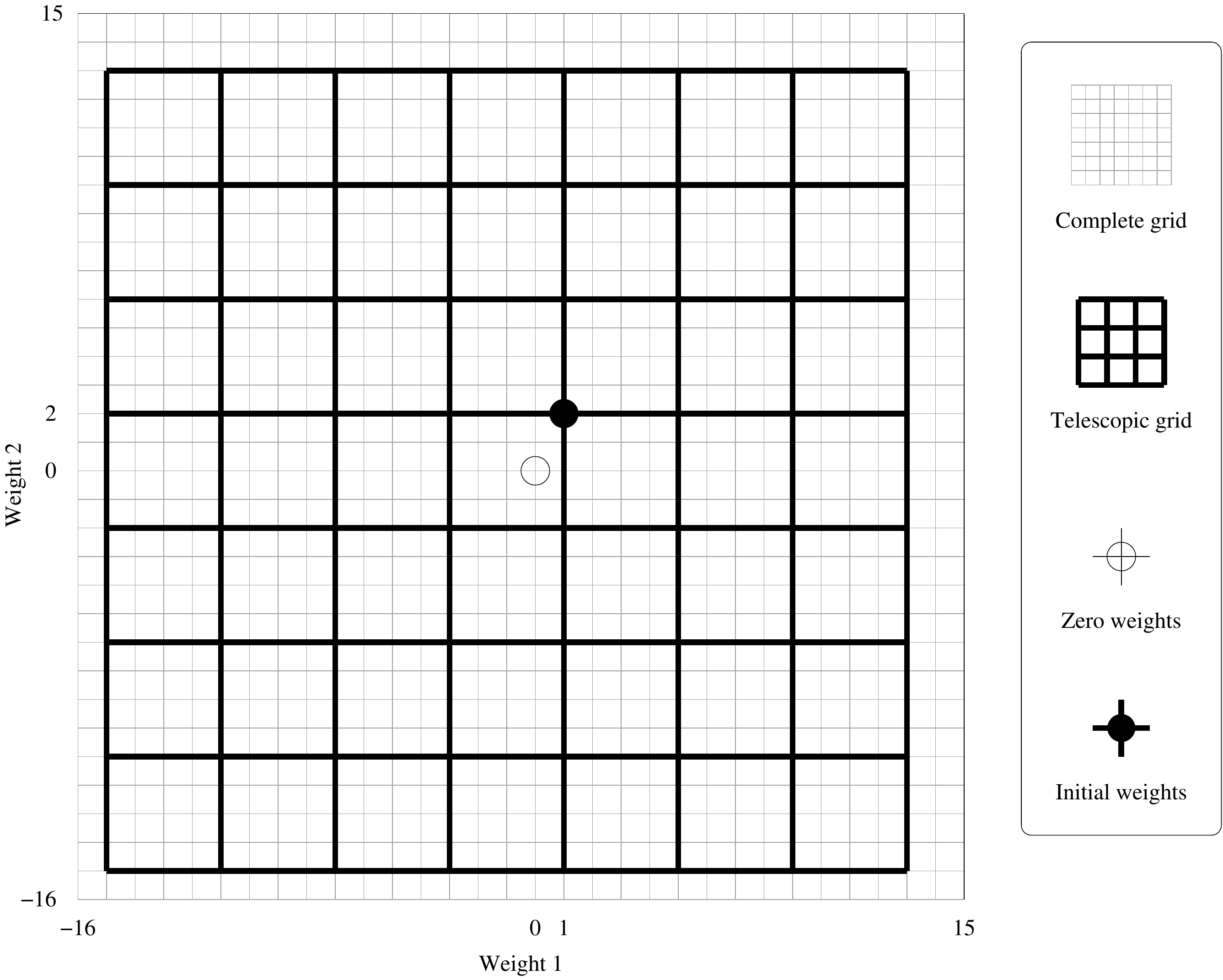}
	\caption{2-weight network, $n=5$ bits per weight, initialized with small values (black dot).
	If only the $n'=2$ most significant bits are changed, the accessible weight set (thick grid)
	does not include $(0,0)$.}
	\label{fig:telegrid}
\end{figure}
\subsubsection{Initialization in telescopic BLM}
Starting from small weights before running telescopic BLM requires a few considerations.
Telescopic BLM initially operates on the most significant part of the weight representation,
leaving the least significant bits untouched, thereby preventing any weight to return to zero
until the final phase when it operates on all bits.\\
Fig.~\ref{fig:telegrid} shows the case where two $n=5$-bit weights are initialized with small values; for simplicity, the
discretization step~(\ref{eq:discretization step}) has been set to $\epsilon=1$.
If the initial step only involves the $n'=2$ most significant bits, then the grid of actually accessible weight values
is offset with respect to the origin, and weights are not allowed to vanish or, in some cases, even become smaller until later phases.\\
In some cases, the small offset acts as a beneficial noise source, forcing small random contributions and encouraging the use of
spontaneous features, in the spirit of reservoir learning. Otherwise, it is possible to alleviate this problem by choosing random initial weights
that are multiple of $\eta=2^{n-n'}\epsilon$ (clearly, $w_\text{init}>\eta$, therefore initial weights are not going to be as small),
so that all least significant bits are forced to zero.

In this paper, for lack of space, we concentrate mostly on the latest two options.


\section{Tests on the two-spirals problem}
\label{sec:two-spirals}

The purpose of the experiments in this section is to test the effect of
the BLM algorithm on a two-dimensional benchmark task, so that
results can be visualized to develop some intuition before the second
series of quantitative tests in the following sections.

\begin{figure}[tbp]
    \includegraphics[width=\hsize]{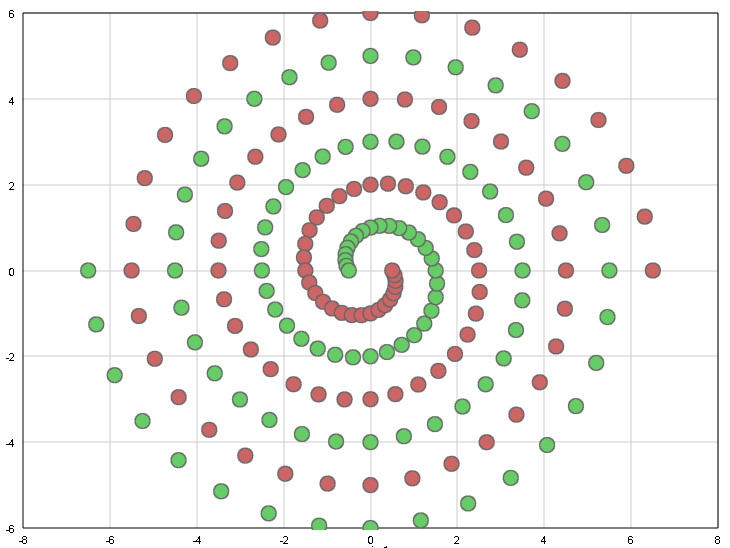}
    \caption{The two-spirals problem.}
    \label{fig:two-spirals}
\end{figure}
A highly nonlinear classification benchmark which is particularly difficult
for gradient descent optimization is the
``two spirals'' problem developed in \cite{lang1988learning}.
The data set is taken from the CMU Neural Networks Benchmarks.
It consists of 193 training instances and 193 test instances located on a 2D surface.
They belong to one of two spirals, as shown in Fig.~\ref{fig:two-spirals}

It is possible to solve this problem with one or two hidden layers,
but architectures with two hidden layers need less connections and can learn faster.
The architecture we consider here is $2-20-20-1$ with bias.
The hidden units consist of symmetric sigmoids (hidden layers) and the usual
$0-1$ sigmoid for the output unit.
All points are used for training, the final mapping is shown with a contour plot
for a visual representation of its generalization abilities.

\begin{figure}[tbp]
    \includegraphics[width=\hsize]{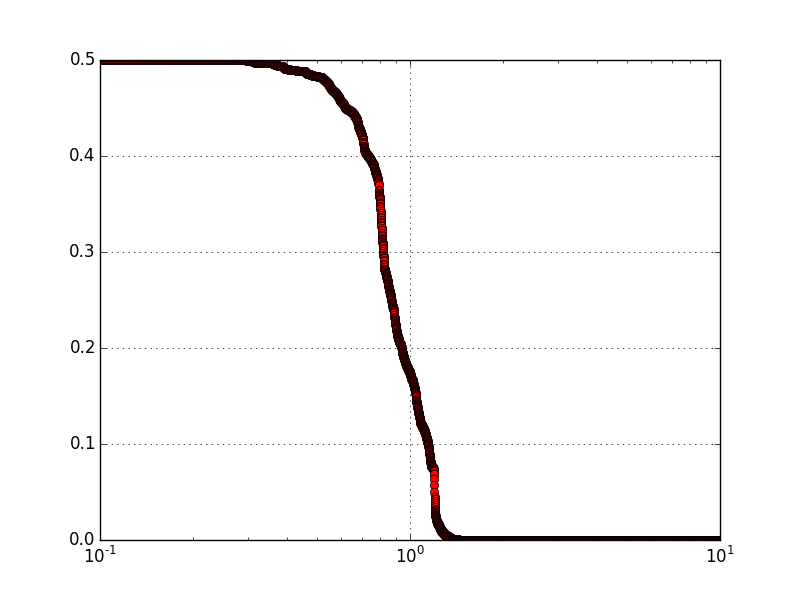}
    \caption{Two-spirals with OSS: RMS error as a function of CPU time. Log time axis.}
    \label{fig:two-spirals-oss-rmse}
\end{figure}
Simple backpropagation tends to remain trapped in local minima or lead to
excessive CPU times.  The one-step secant (OSS) technique\cite{Bat92,BaTe94c} based
on approximations of second derivatives from gradients
is much more effective and is among the state-of-the-art methods for training
this kind of highly nonlinear systems.
The RMS error of OSS as a function of CPU time is shown in Fig. \ref{fig:two-spirals-oss-rmse}.
It can be shown that a large number of OSS iterations (about~2000) are spent in the initial
phase without significantly improving the RMS error, then the proper path in weight space
is identified and the error rapidly goes to zero.

\begin{figure}[tbp]
    \includegraphics[width=\hsize]{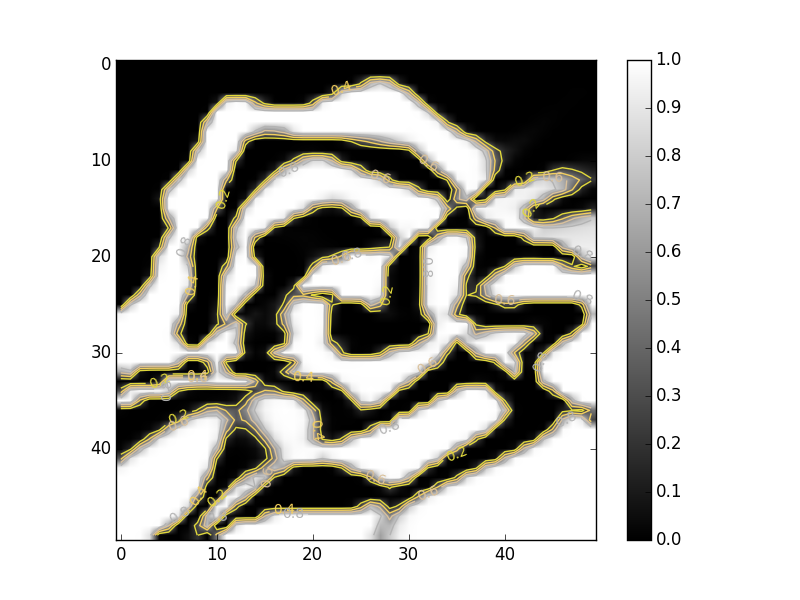}
    \includegraphics[width=\hsize]{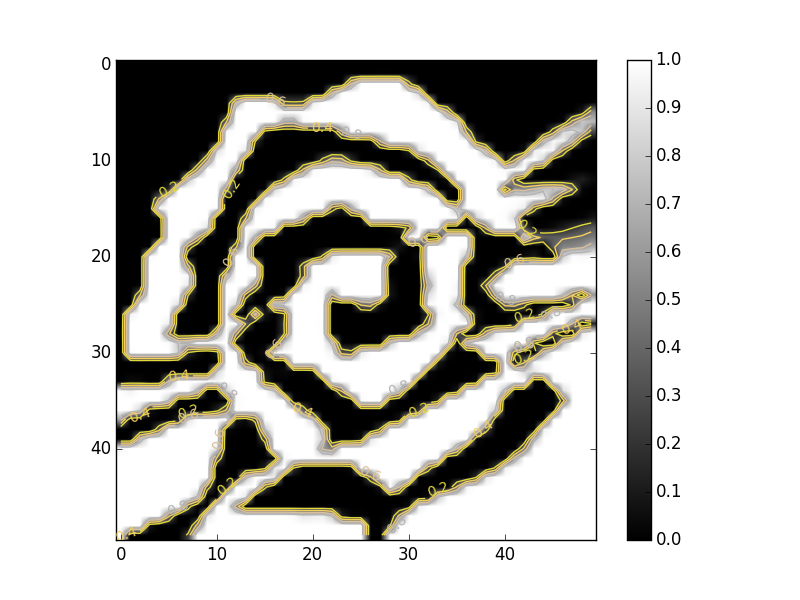}
    \caption{Two-spirals with OSS: intermediate result (above, 2000 iterations, RMS 0.02) and final map obtained.}
    \label{fig:two-spirals-oss}
\end{figure}
The final results obtained with OSS are shown in Fig. \ref{fig:two-spirals-oss}.
Some signs of over-training are clearly visible as stripes in the map.
Early stopping does not help. Let's focus on the final map for a fair comparison with those obtained
by BLM at the local minimum.

\begin{figure}[tbp]
    \includegraphics[width=\hsize]{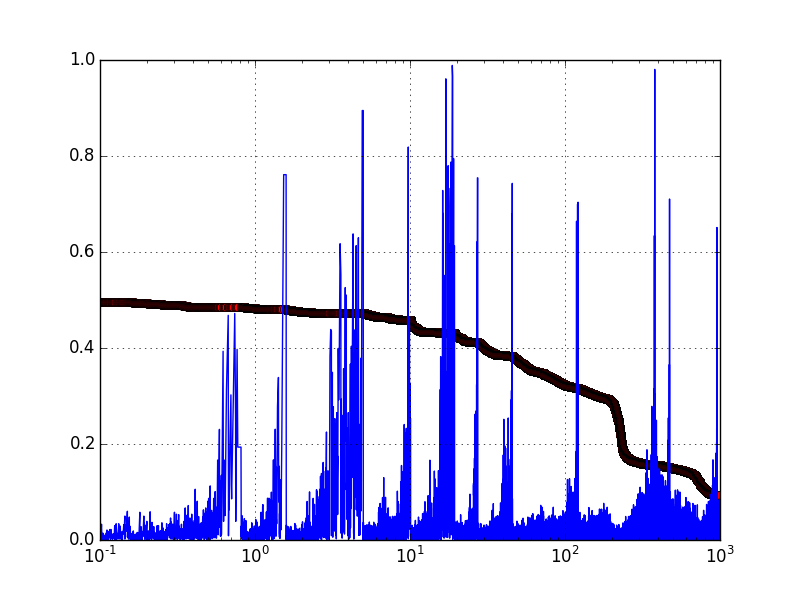}
    \caption{Two-spirals with telescopic BLM, 20-20 architecture, training RMSE as a function of CPU time,
	one curve (top, red) shows the RMSE evolution, the second curve (blue, bottom)
    shows the percentage of the neighborhood considered at each step.}
    \label{fig:two-spirals-tele-rmse}
\end{figure}
Results for telescopic BLM are in Fig. \ref{fig:two-spirals-tele-rmse}
The architecture is the same as that used for OSS.
The specific parameters of BLM are:
		initial weight range     .001,
		initial number of bits	2,
		12 bits,
		weight range     6.0.
		
It can be observed that the percentage of the neighborhood considered at each step
is below 1\% for most steps, with rare jumps up to 5-10\% (these rare jumps
are visible in the plot which shows all 726702 steps executed in 1000 seconds).
The fraction of the neighborhood explored by first-improve tends to grow
only in the final part of search for each number of bits, when the local
minimum is close and more and more neighbors need to be examined before
finding an improving move. After the local minimum is reached, a new bit
is added to each weight so that the fraction drops again to very low values.
Each macroscopic spike therefore corresponds to the moment when a local minimum
is reached and the number of bits is increased.

\begin{figure}[tbp]
    \includegraphics[width=0.32\hsize]{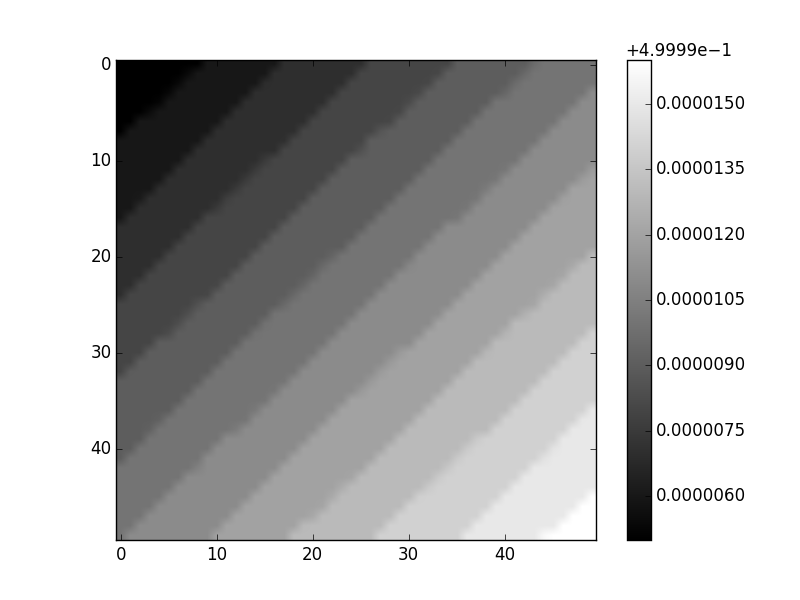}
    \includegraphics[width=0.32\hsize]{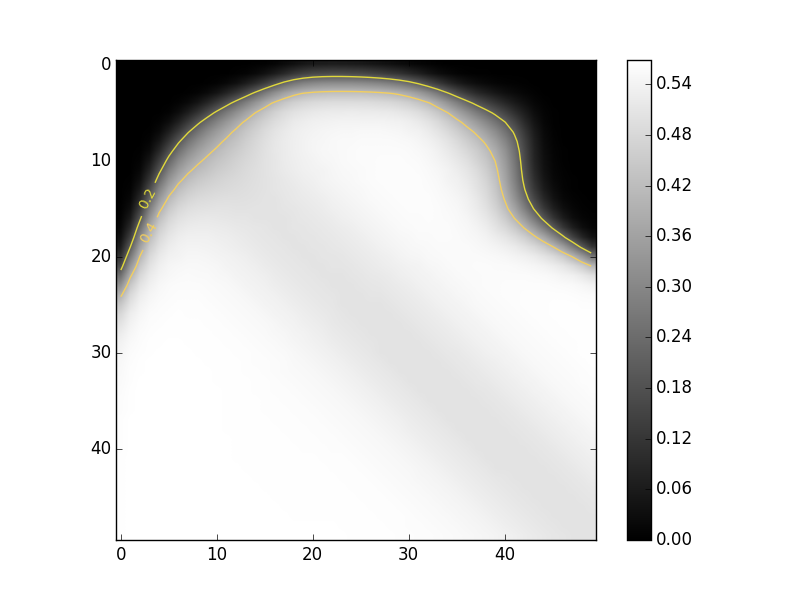}
    \includegraphics[width=0.32\hsize]{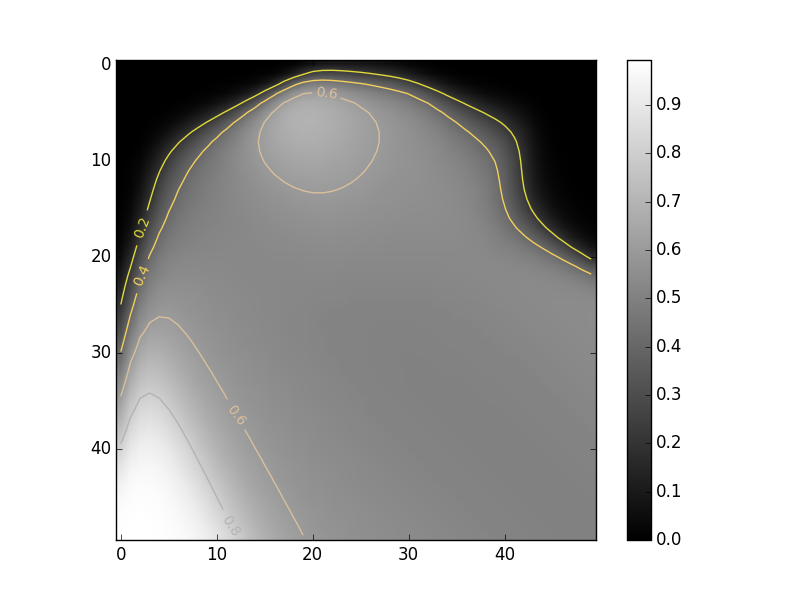}
    \includegraphics[width=0.32\hsize]{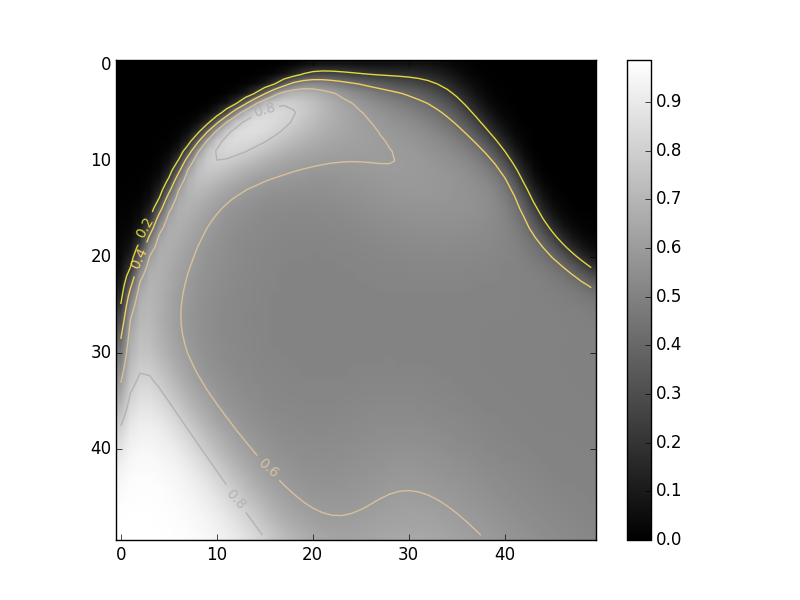}
    \includegraphics[width=0.32\hsize]{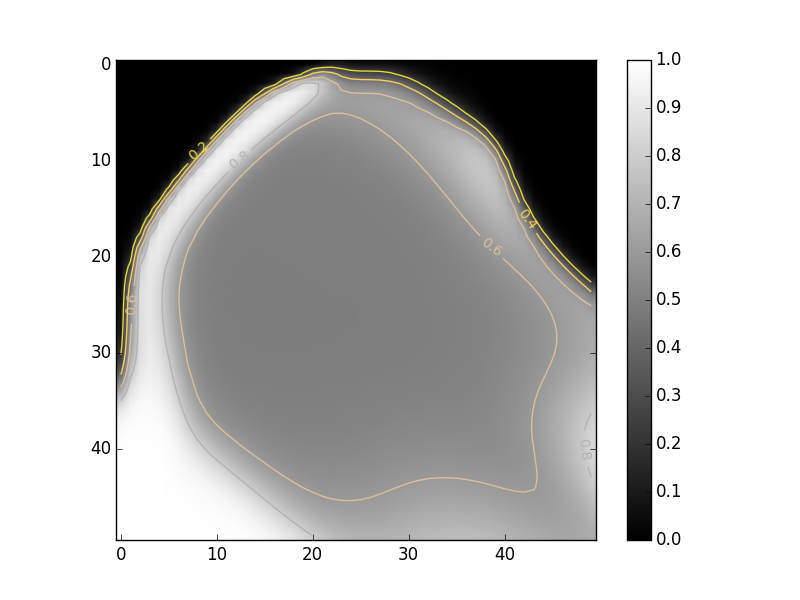}
    \includegraphics[width=0.32\hsize]{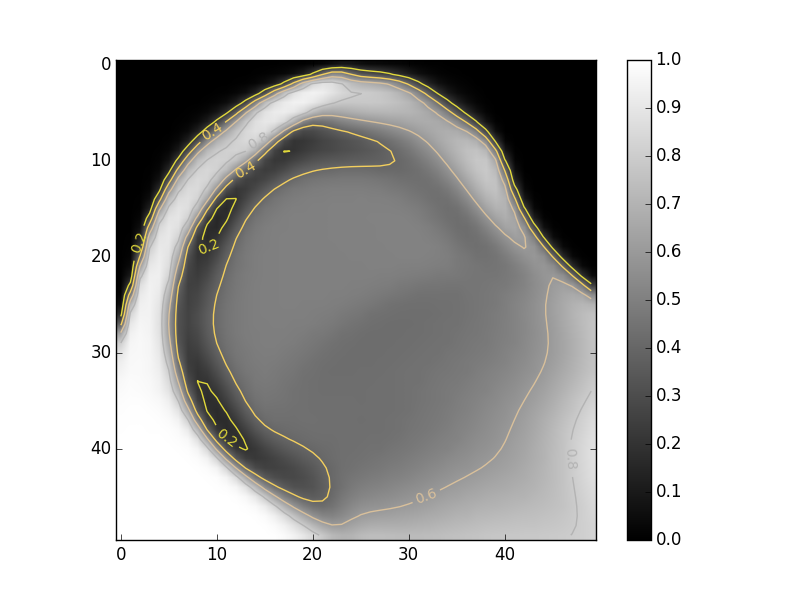}
    \includegraphics[width=0.32\hsize]{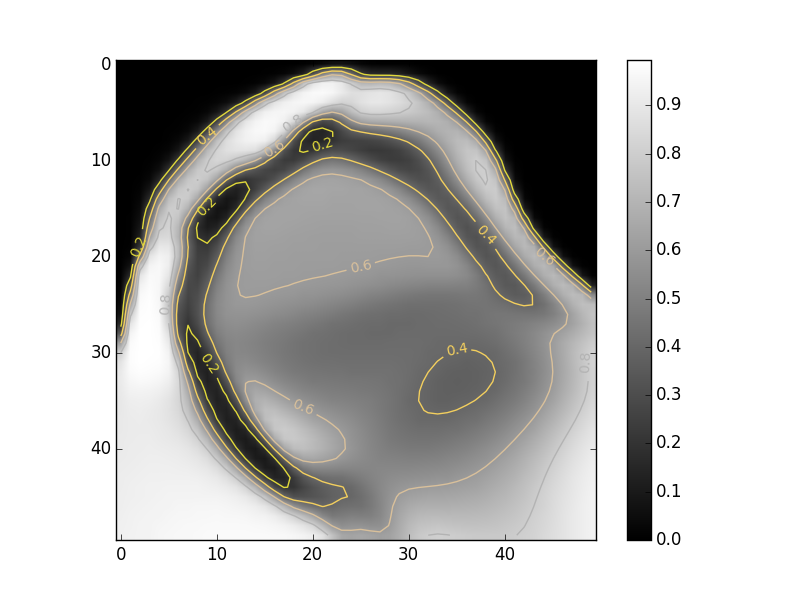}
    \includegraphics[width=0.32\hsize]{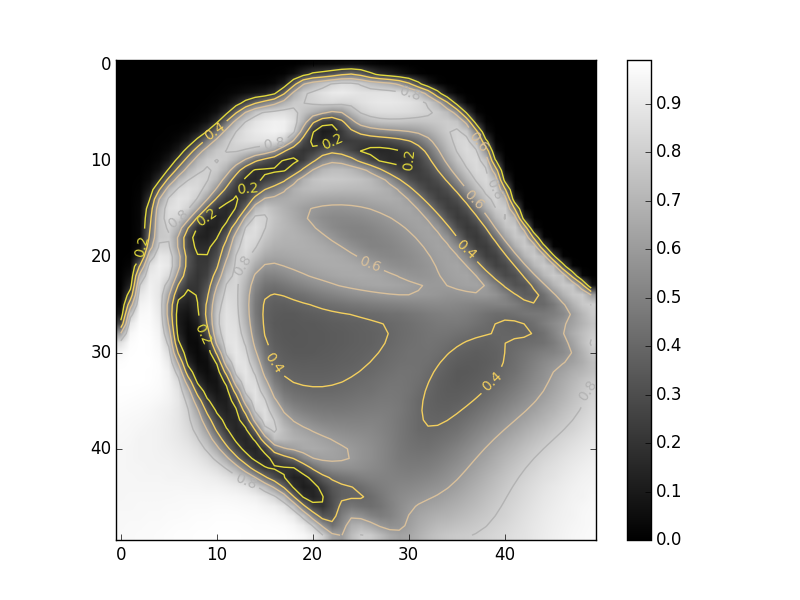}
    \includegraphics[width=0.32\hsize]{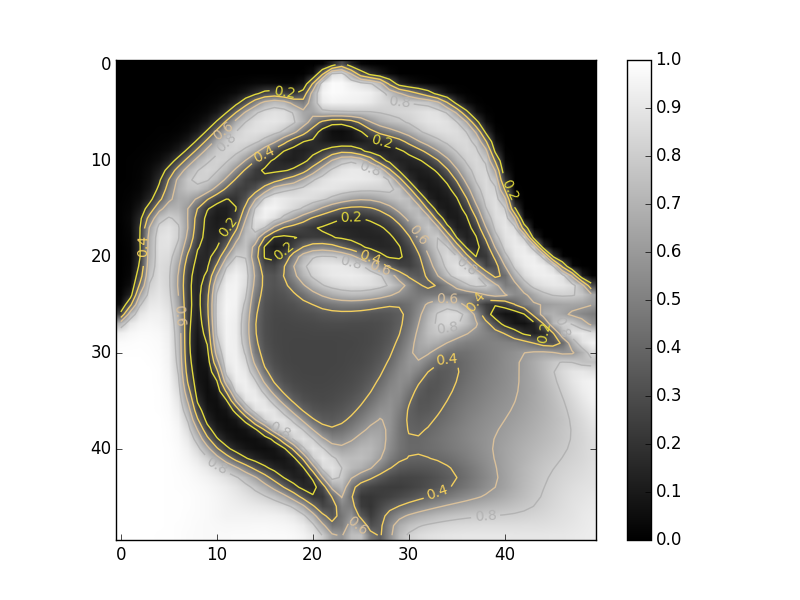}
    \includegraphics[width=0.32\hsize]{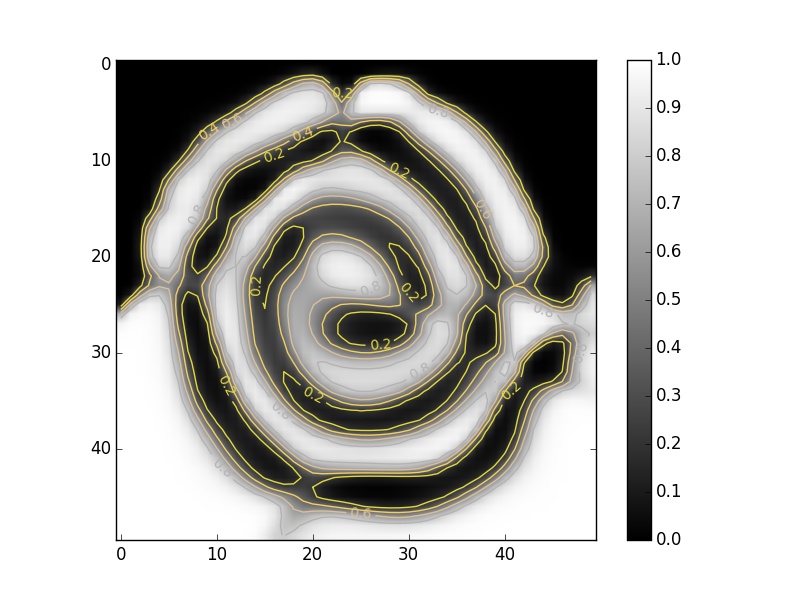}
    \includegraphics[width=0.32\hsize]{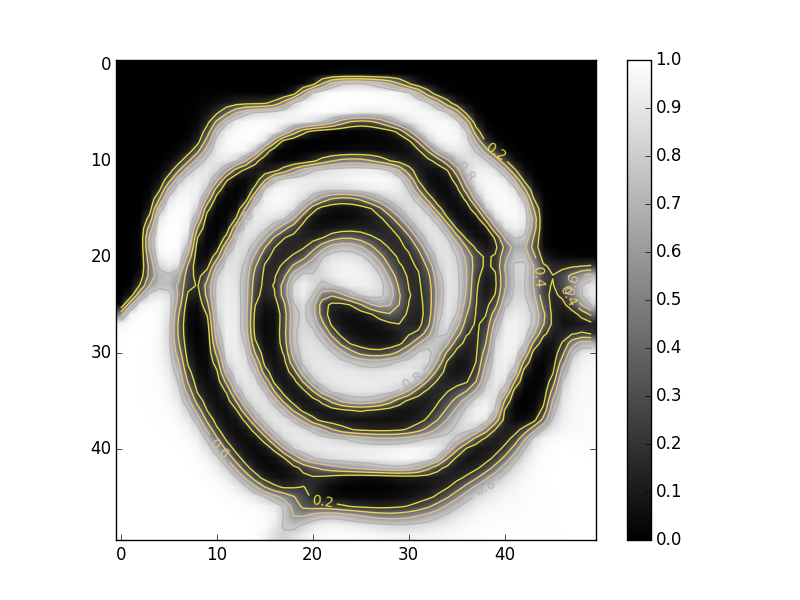}
    \includegraphics[width=0.32\hsize]{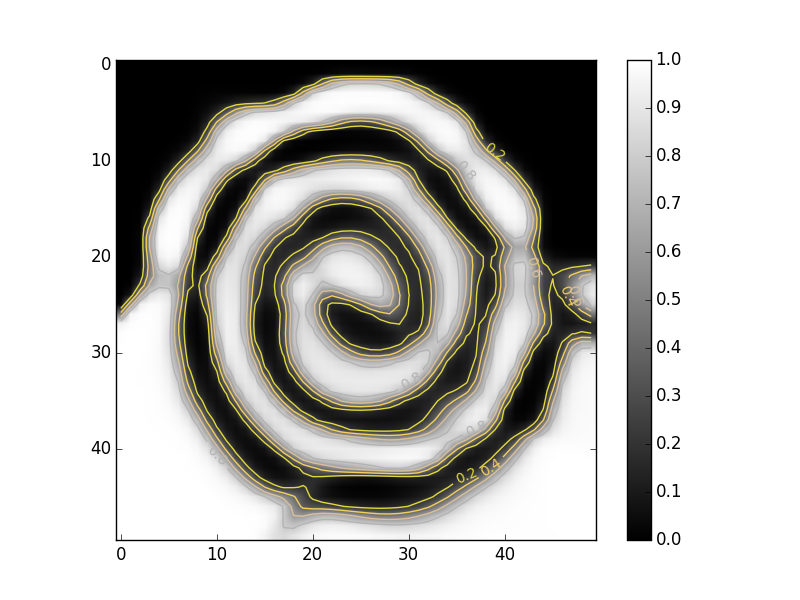}
    \caption{Two-spirals with telescopic BLM, 20-20 architecture.
	The capability of the map to represent the spiral increases when
	the number of bits per weight increases (from 1, top left, to 12, bottom right).}
    \label{fig:two-spirals-tele}
\end{figure}

The various local minima correspond to the mappings of Fig. \ref{fig:two-spirals-tele}.
The telescopic ``multi-scale'' algorithm works by first modelling the overall
large-scale circular structure and the south positive and north negative area.
Then finer and finer internal details are fixed. The spiral maps is already well
formed with 10 bits, while the two additional bits add some fine-tuning.


\begin{figure}[tbp]
   \includegraphics[width=1.0\hsize]{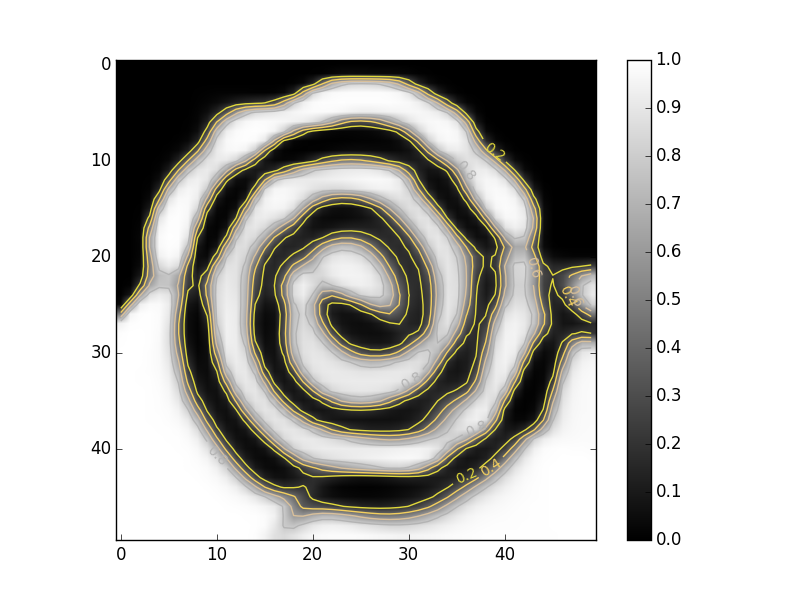}
    \caption{Two-spirals with telescopic BLM, 20-20 architecture, final mapping obtained.}
    \label{fig:two-spirals-evensmoother}
\end{figure}

The final mapping obtained (Fig. \ref{fig:two-spirals-evensmoother})
is very smooth, without the signs of overtraining
which are evident in the mapping for OSS. This result is surprising given that is
is obtained with discretized weights represented with 12 bits.

%

Because BLM acts by local search changing individual bits, it is
simple to consider variations of the algorithms.
For example, \emph{sparsity} can be encouraged, by aiming at networks
in which a large number of weights is fixed to zero (and therefore
can be eliminated from the final realization).

To encourage sparsity, in the following test only 50\% of the weights
are initialized with values different from zero, and the analysis of potential moves in the
neighborhood at each step is changed as follows. First all bits in \emph{non-zero} weights  are tested for a possible change.
Only of no such bit is identified, bits in zero weights are considered.
The philosophy is ``Let the current neurons fully express their potential
before adding additional neurons (with non-zero weights)''.

\begin{figure}[tbp]
   \includegraphics[width=1.0\hsize]{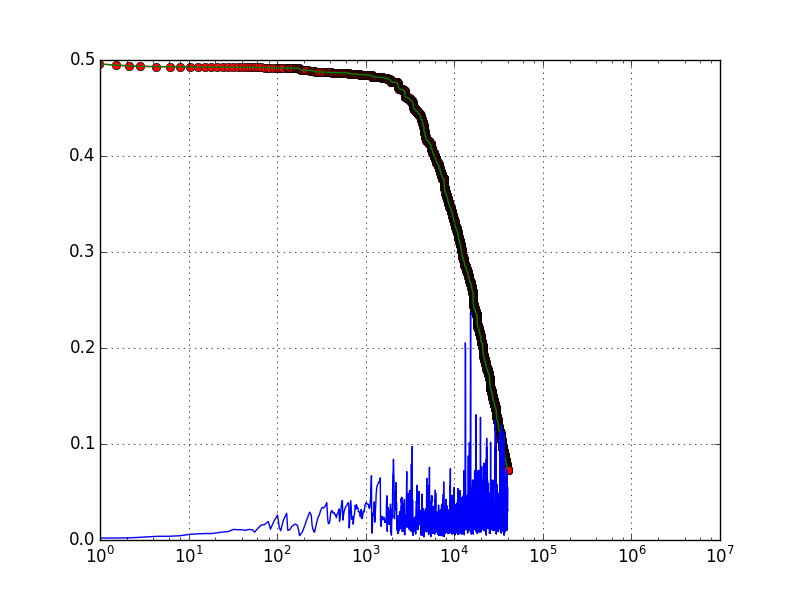}\\
   \vskip-.7\hsize
   \hskip.6\hsize
   \includegraphics[width=.3\hsize]{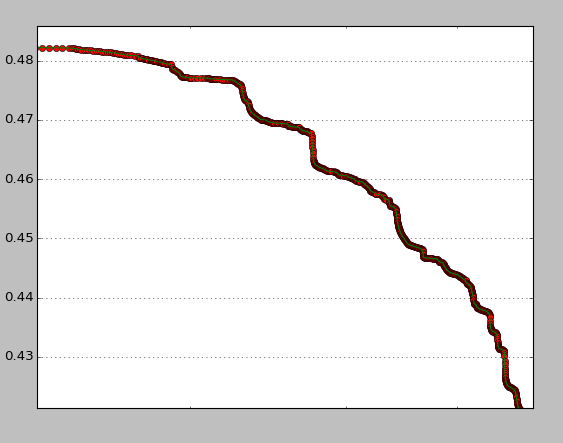}\\
   \vskip .33\hsize
    \caption{Two-spirals with sparsity enforcing, evolution of training (Circles: RMSE as a function of CPU time; line: fraction of explored neighborhood). Inset: zoom-in of the initial improvement; the step-like behavior is due to new neurons becoming active
    (new non-zero weights).}
    \label{fig:two-sparse-training}
\end{figure}

\begin{figure}[tbp]
   \includegraphics[width=1.0\hsize]{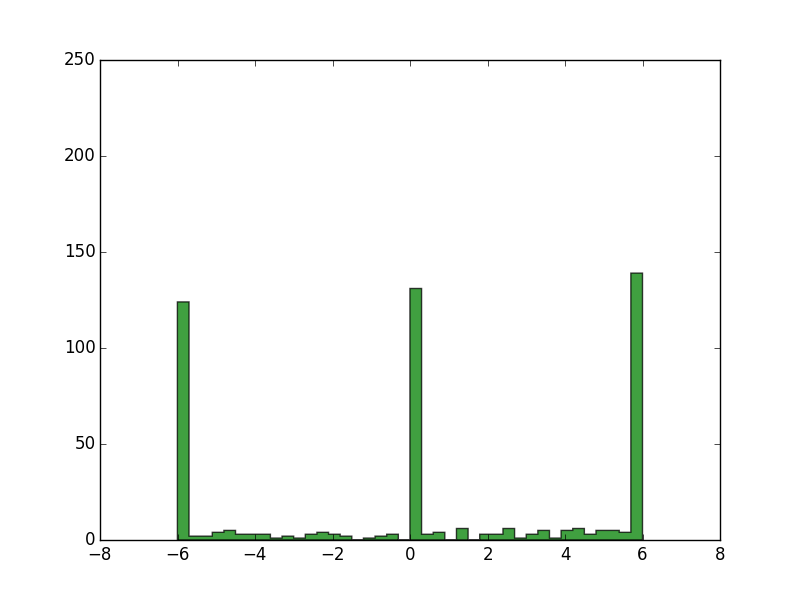}
    \caption{Two-spirals with sparsity enforcing, final distribution of weights.}
    \label{fig:two-sparse-wei}
\end{figure}

\begin{figure}[tbp]
   \includegraphics[width=1.0\hsize]{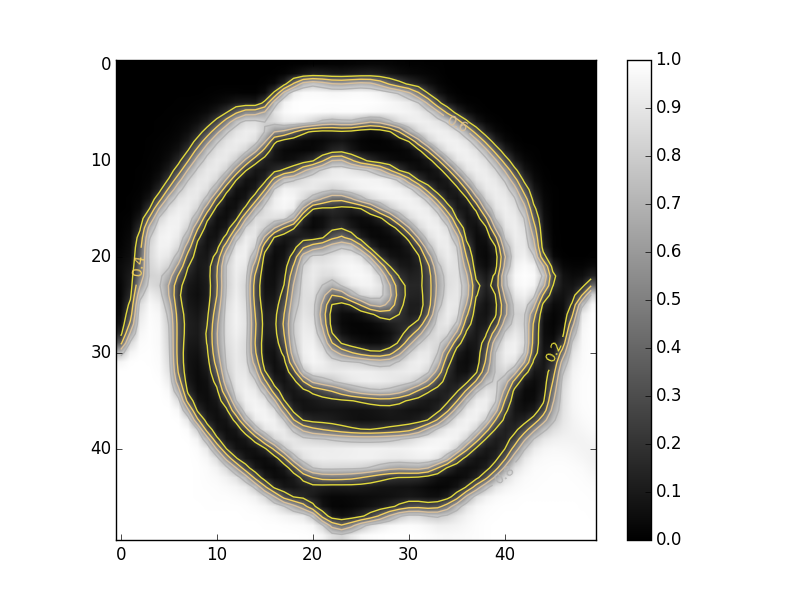}
    \caption{Two-spirals with sparsity enforcing, final mapping obtained.}
    \label{fig:two-sparse-final}
\end{figure}

The evolution of this sparsity-enforcing training is shown in Fig. \ref{fig:two-sparse-training}.
It can be observed that the final distribution of weights is indeed sparse (Fig. \ref{fig:two-sparse-wei}),
with a peak at zero and two peaks at the smallest and largest possible values. 
The final mapping is smoother than the one obtained without enforcing sparsity (Fig. \ref{fig:two-sparse-final}).

While we have no space in this paper to investigate the sparsity issues
for all benchmarks, the above preliminary results are encouraging
and we plan to extend the analysis in future work.


\section{Tests on a control problem: the inverted pendulum}
\label{sec:pendulum}

\begin{figure}[tbp]
	\includegraphics[width=\hsize]{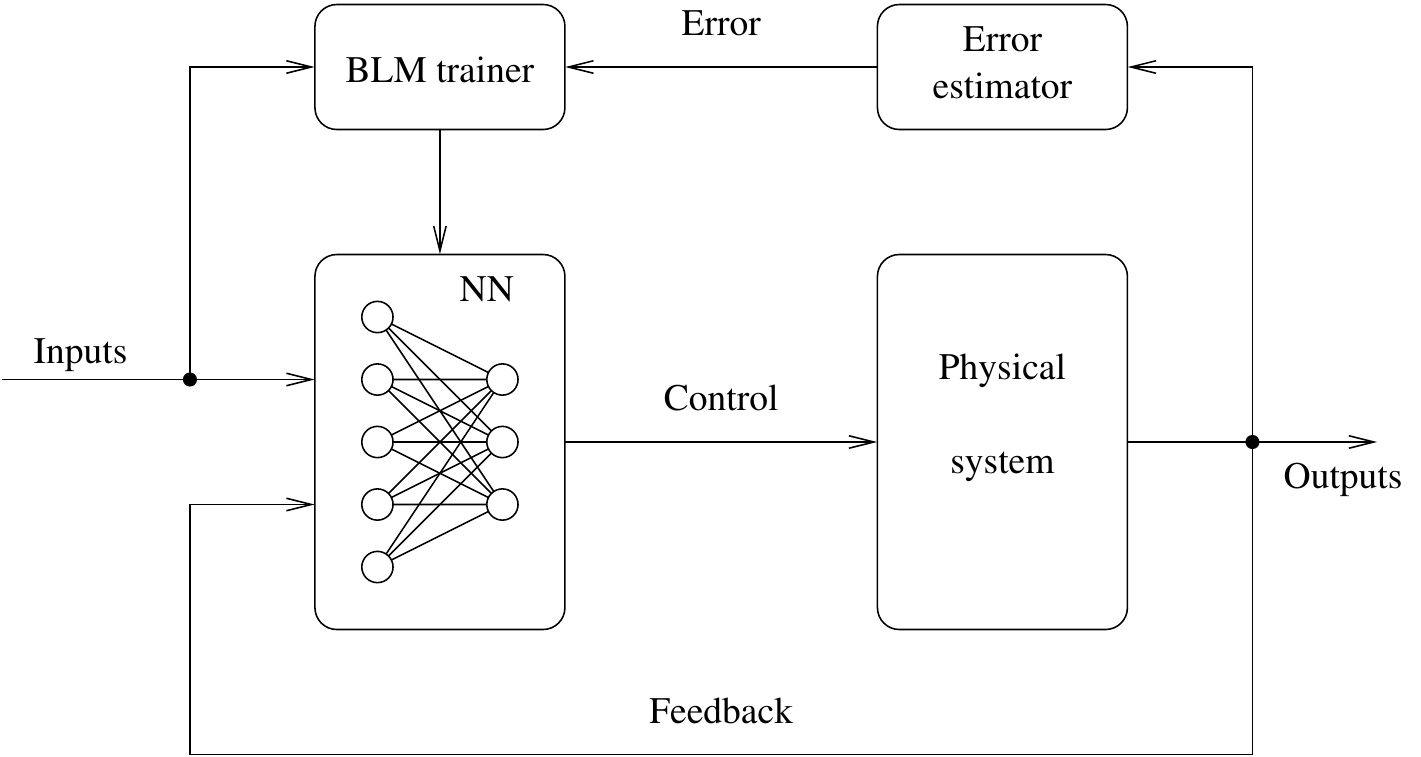}
	\caption{BLM-trained neural network as a controller.}
	\label{fig:BLM controller}
\end{figure}
In this Section we describe our investigation on the use of BLM-trained networks for applications in
control problems, with the network acting as a controller component of a feedback loop, as shown
in Fig.~\ref{fig:BLM controller}.

In the scheme, we can exploit the main advantage of BLM as a derivative-free training algorithm: the controlled physical system
can be treated as a black box, with the only obvious addition of a system-specific error evaluation function (top right) whose
numeric output measures the correctness of the system's behavior.

\begin{figure}[tbp]
	\begin{minipage}{.6\hsize}
		\includegraphics[width=\hsize]{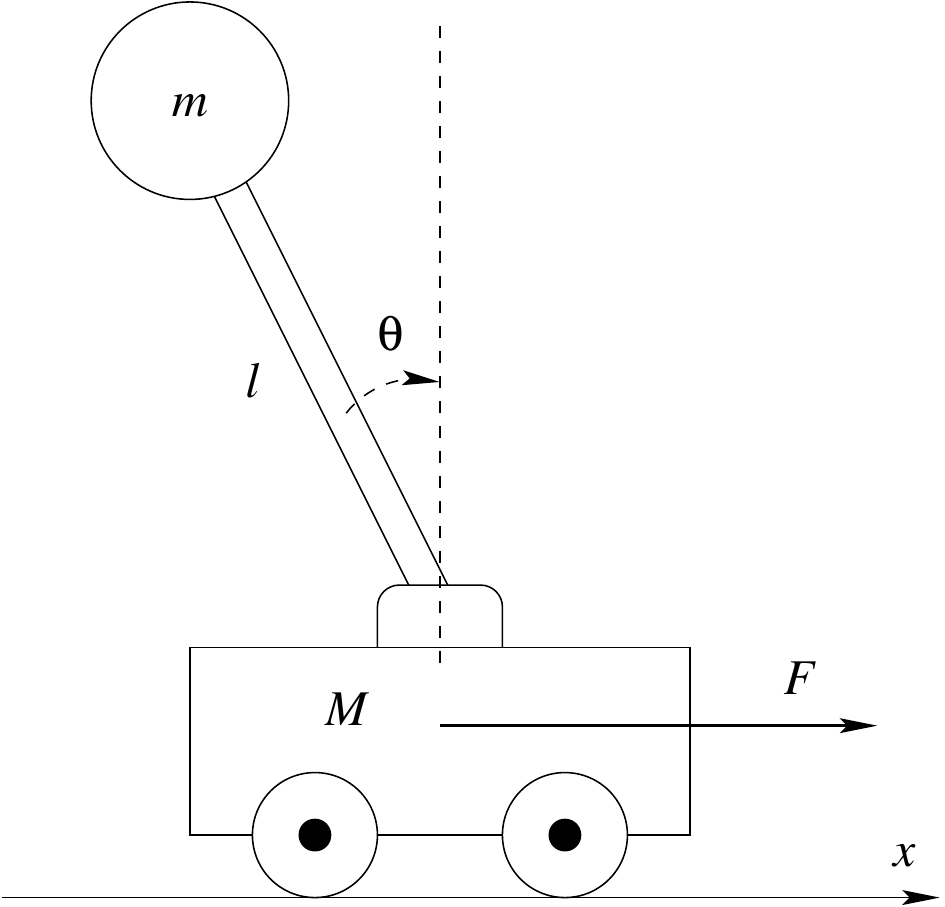}
	\end{minipage}
	\hfill
	\begin{minipage}{.35\hsize}
		\noindent Parameters:\\
			$M = m = 1\text{kg}$,\\
			$l = 1\text{m}$.\\
		\null\\
		Time discretization:\\
			$t \in [0,100\text{s}]$,\\
			$\Delta t = .01\text{s}$.\\
		\null\\
		Initial conditions:\\
			$x_0 = 0$,\\
			$\theta_0 \in [-.4,.4]$.
	\end{minipage}
	\caption{Inverted pendulum on a cart.}
	\label{fig:pendulum}
\end{figure}
In our case study, we simulate the well-known system shown in Fig.~\ref{fig:pendulum}:
an inverted pendulum mounted on a cart on which a control force is applied. The cart has mass $M$
and is constrained to move along direction $x$, while the pendulum has length $l$ and mass $m$
applied on the end. Other masses are negligible. The control variable is a force applied
to the cart along its moving direction.

The dynamics of the system are described by the following second-order, nonlinear system
of differential equations:
\begin{eqnarray*}
\ddot x &=& \frac{F-m\sin\theta(l\dot\theta^2-g\cos\theta)}{M+m\sin^2\theta}\\
\ddot\theta &=& \frac{\ddot x\cos\theta+g\sin\theta}l.
\end{eqnarray*}

The correct behavior of the system is to keep the pendulum in upright position $\theta=0$, with the cart
at a specified coordinate $x=0$. If we simulate the system in the time interval $[0,T]$, then the error
is given by a combination of the mean squares of the two position variables $x,\theta$:
\begin{equation}
	\label{eq:dynamic error}
	\text{Err}(x,\theta)=\frac1{T-t_\text{min}} \int_{t_\text{min}}^T\left(\theta^2(t)+\lambda x^2(t)\right)\de t,
\end{equation}
where $\lambda$ is a suitable weighting factor (in our experiments, $\lambda=0.01\text{m}^{-2}$),
while $t_\text{min}\ge0$ defines a transient portion of the time interval that does not contribute to the error.

\subsection{Complete feedback}

In our first test, we use a feed-forward neural network with a 5-neuron hidden layer (with $\tanh$ as transfer function)
and a single output neuron that provides the control force. The network has four inputs: the two position variables $x,\theta$
and their time derivatives $\dot x,\dot\theta$. We use 16-bit weights, with $10$ as maximum value,
initialized to small random values in the $[-10^{-2},10^{-2}]$ interval.

For the training procedure, we generate $50$ pendulum simulations, each running for a simulated time $T=100\text s$.
We use first-order
(Euler) integration with time increment $\Delta t=10^{-2}\text{s}$. The control force is kept constant for $10$ consecutive steps,
after which the system's status is fed to the neural net, and the new output is used for the subsequent 10 iterations.
Therefore, the net is queried $1000$ times per simulation. The first $10$ network queries, corresponding to $t_\text{min}=1\text{s}$
of simulated time, don't account for the error.
Every $100$ training iterations, a different set of $50$ simulations is used to validate the training and
the weights corresponding to the best validation are saved.

\begin{figure}[ptb]
	\centering
	\includegraphics[width=\hsize]{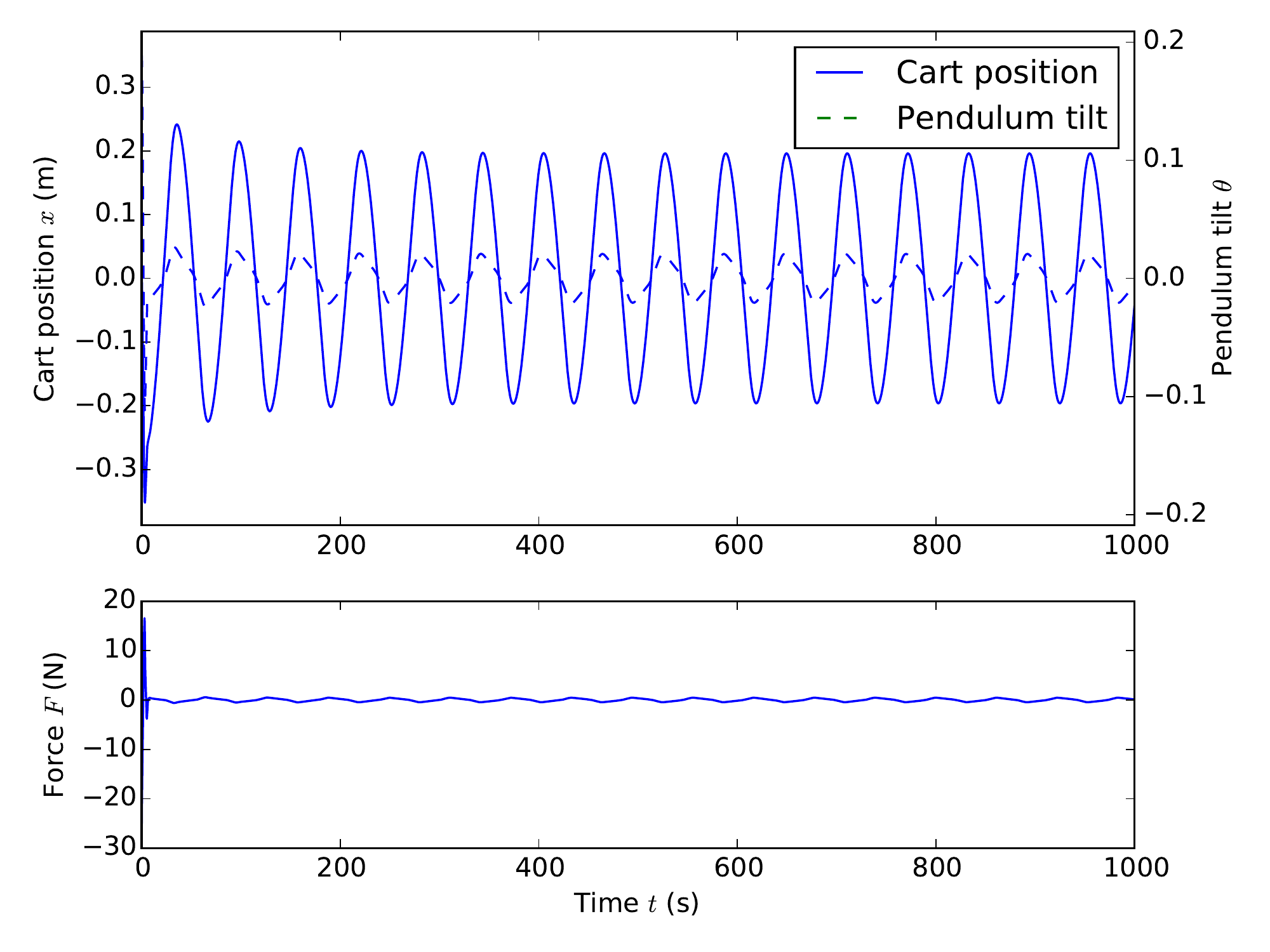}
	\includegraphics[width=.75\hsize]{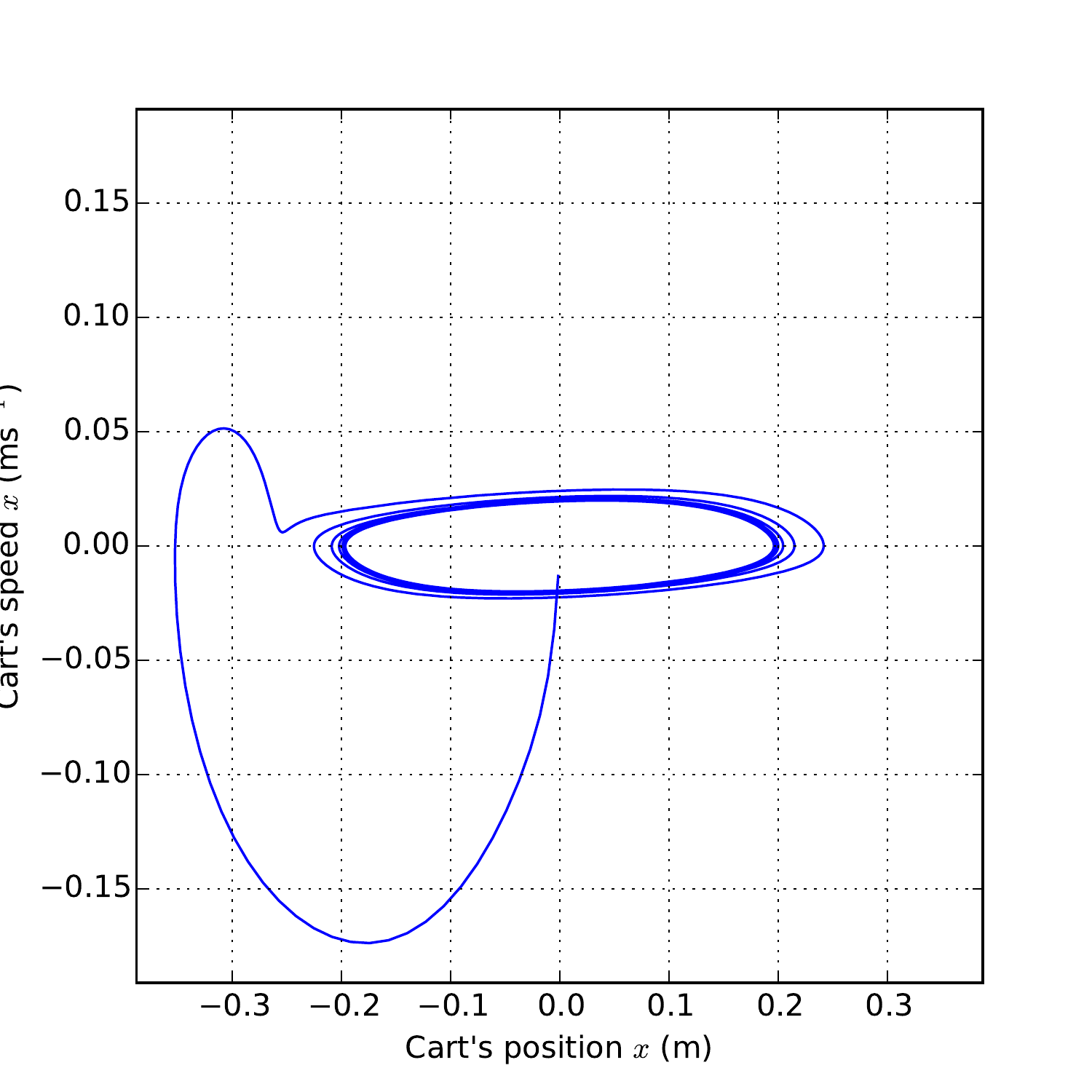}
	\includegraphics[width=.75\hsize]{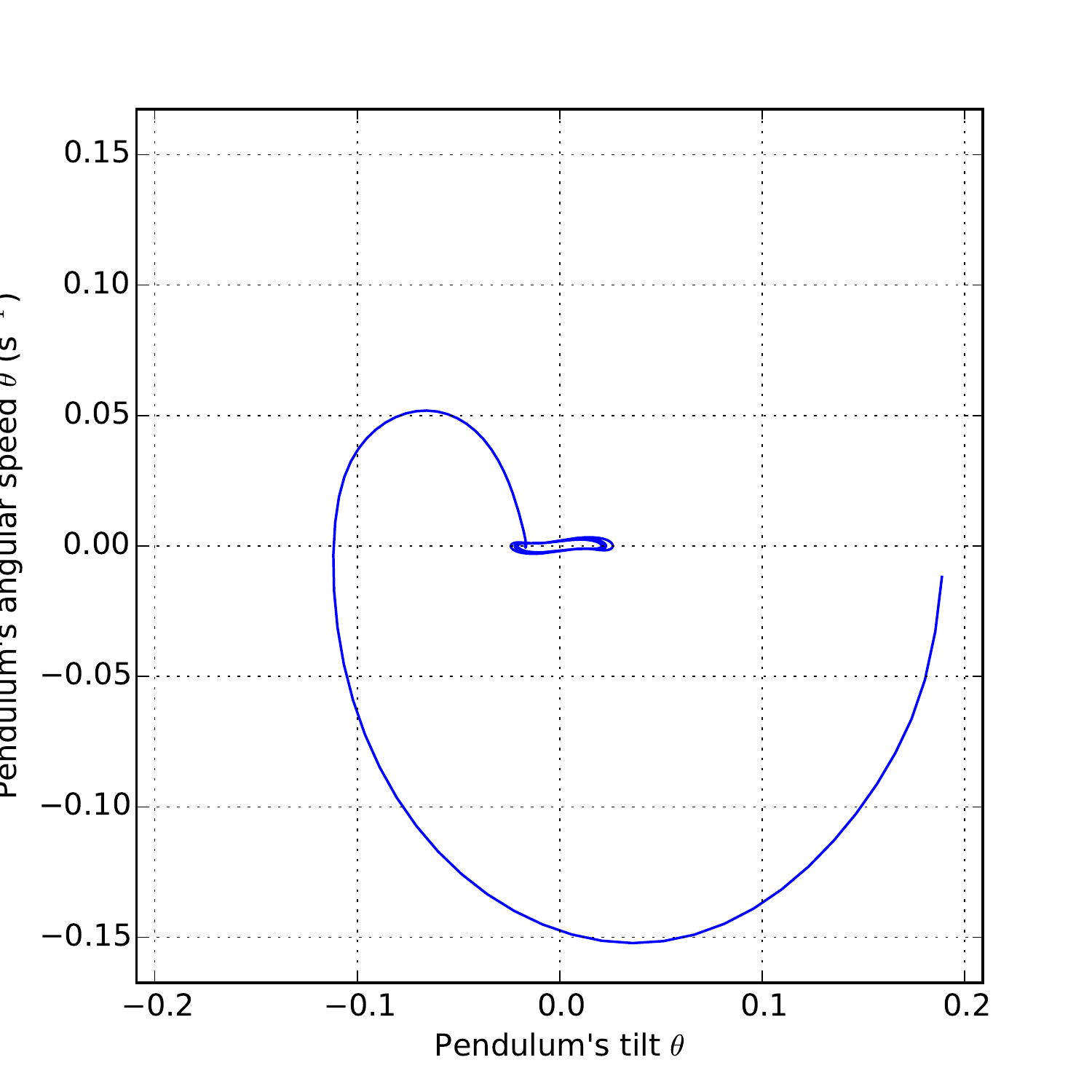}
	\caption{Inverted pendulum with complete feedback ($x,\theta,\dot x,\dot\theta$), feed-forward MLP
		without recurrence, 5 hidden units and optimization on 2 bits per weight. Two bits are sufficient
		to keep the pendulum from falling, although control is coarse and both the pendulum and the cart keep moving back and forth
		along a limit cycle, as shown by the two phase-space diagrams in the bottom part.}
	\label{fig:inverted complete 2bit}
\end{figure}
Fig.~\ref{fig:inverted complete 2bit} shows the behavior of a network with 5 hidden
units trained with only 2~bits per weight on a sample
test instance; to make sure that the stability is not temporary, the test simulation runs for $1000$ simulated seconds, i.e., tenfold the actual
training and validation period. The top chart of Fig.~\ref{fig:inverted complete 2bit} shows the behavior of the two positional variables $x,\theta$
during the simulation. The network's control, however coarse due to the few representable weight values, is sufficiently precise to prevent the pendulum from falling,
although the cart tends to drift back and forth (remember that we gave a relatively small value to the penalty $\lambda$
associated to the cart position $x$). The middle chart in the same Figure shows the corresponding applied force: after a transient
high force to correct the unbalanced initial conditions of the system, much less force needs to be applied to keep the system upright.
The ensuing periodicity is apparent in phase space, where the pairs of coordinates $(x,\dot x)$ and $(\theta,\dot\theta)$ both
converge to a limit cycle. The mean error of the network as defined in~(\ref{eq:dynamic error}),
computed on $50$ random initial conditions, is $\text{Err}=2.34\cdot10^{-2}$.
\begin{figure}[ptb]
	\includegraphics[width=\hsize]{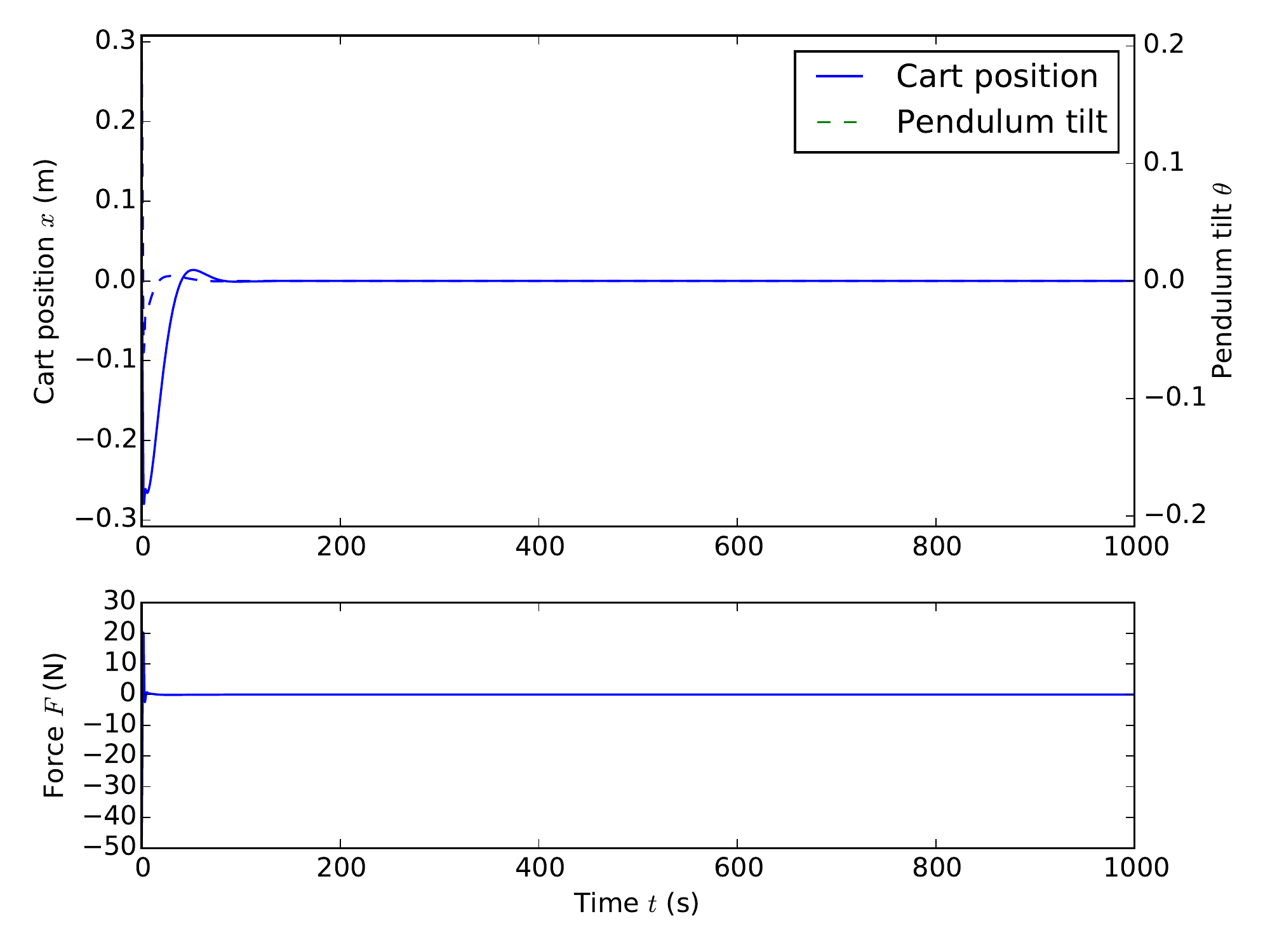}
	\caption{Inverted pendulum with complete feedback ($x,\theta,\dot x,\dot\theta$), feed-forward MLP
		without recurrence, 5 hidden units and telescopic optimization on 16 bits per weight.
		The pendulum is now stabilized within a short time span.}
	\label{fig:inverted complete 16bit}
\end{figure}
If a finer representation is chosen for weights (e.g., 16 bits), then the system comes to a relatively smooth halt
in a short time, as shown in Fig.~\ref{fig:inverted complete 16bit}, with error $\text{Err}=8.9\cdot10^{-3}$.

\subsection{Derivative-free feedback}

Moving to a harder version of the problem, we assume that only the positional variables $x,\theta$ are fed back to the network,
while their time derivatives are not. The resulting network only has two inputs and one output.

In this case, a pure feed-forward network is unaware of the current
inertial status, and the best that we can obtain is a slowly divergent system in which oscillations become larger and larger
while the cart moves away from the central position.

The addition of recurrent connections from the hidden layer to itself, together with a convenient increase of
the number of hidden units, provides the small amount of short-term memory that
the network needs in order not to diverge.

\begin{figure}[ptb]
	\centering
	\includegraphics[width=\hsize]{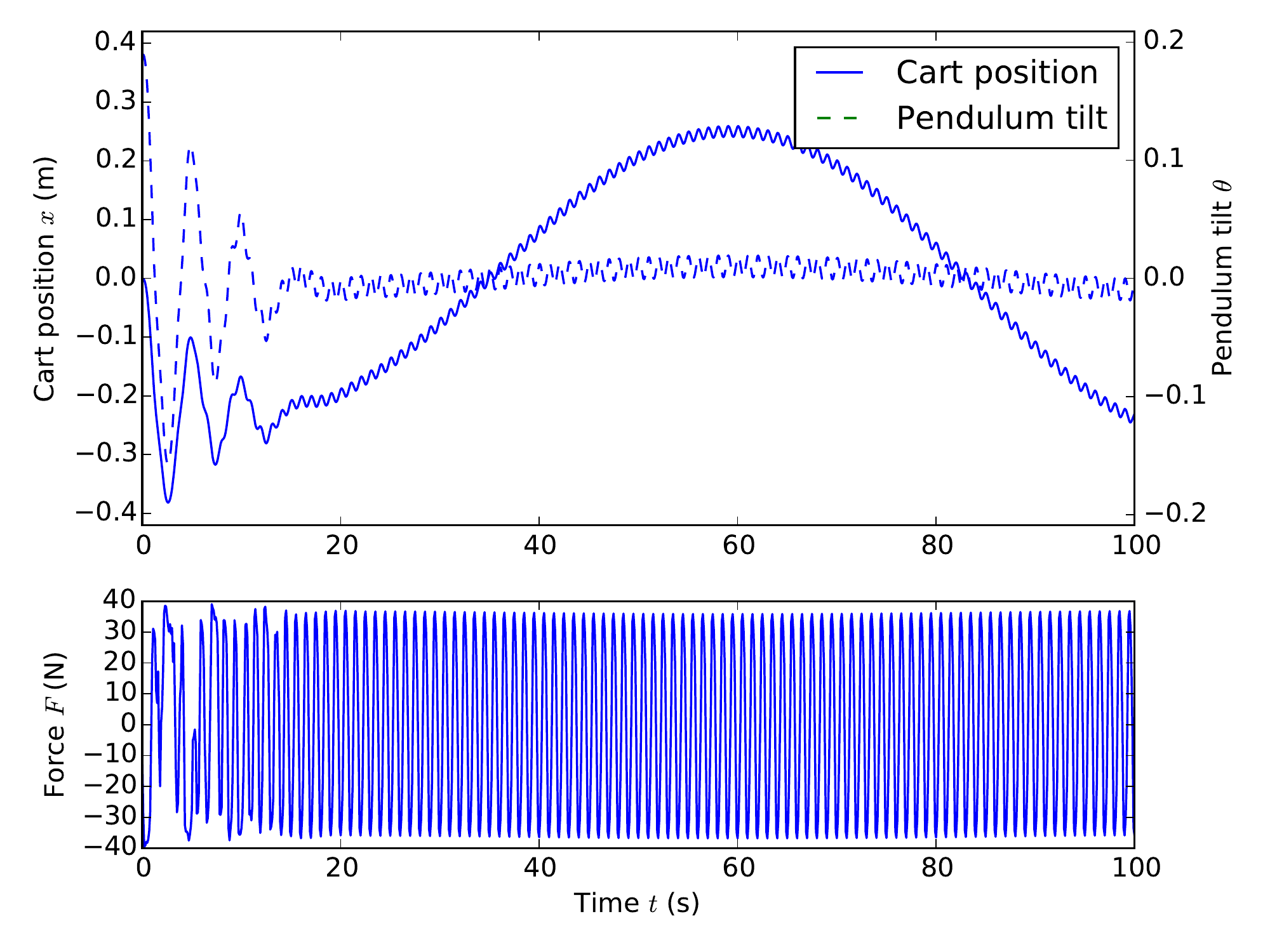}
	\includegraphics[width=.75\hsize]{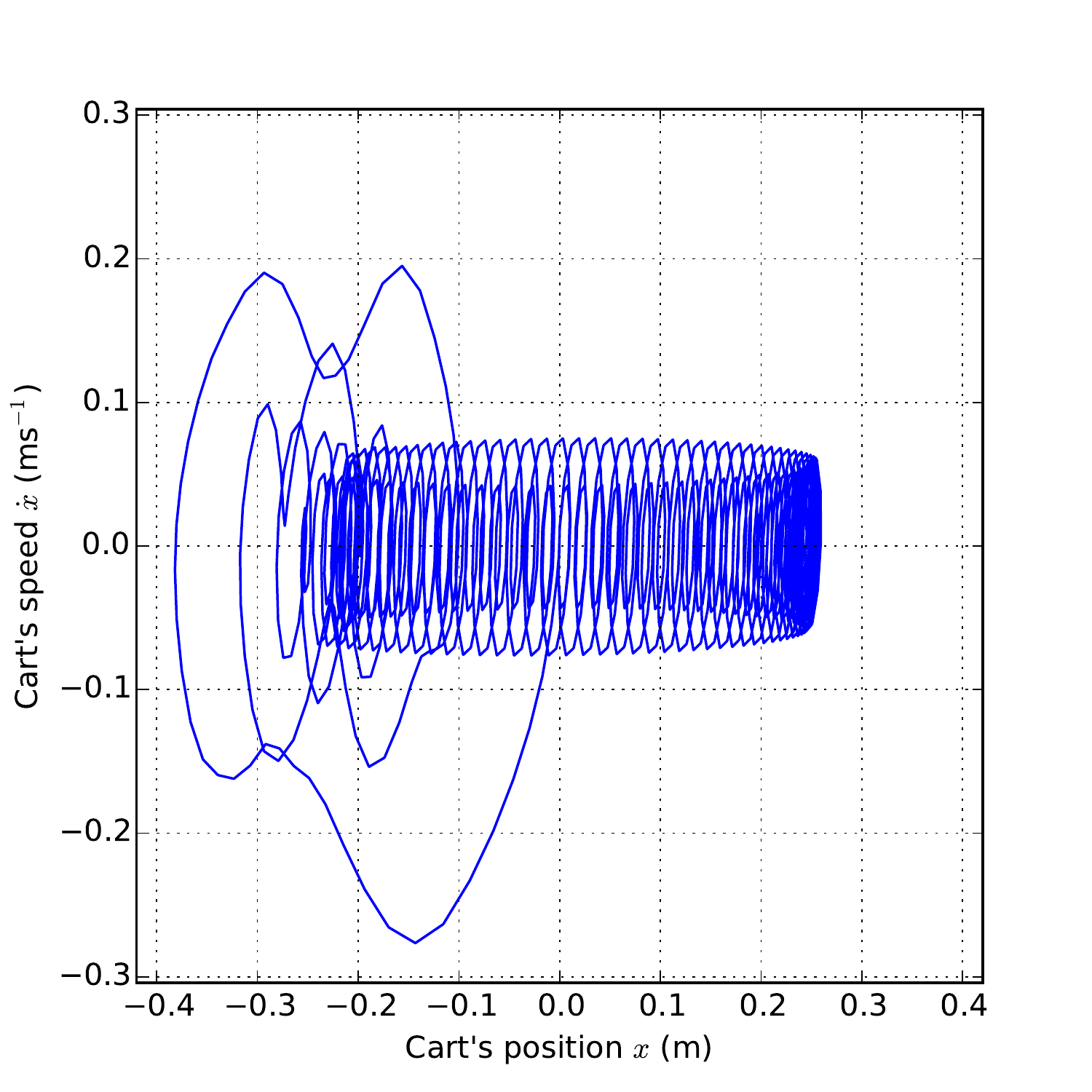}
	\includegraphics[width=.75\hsize]{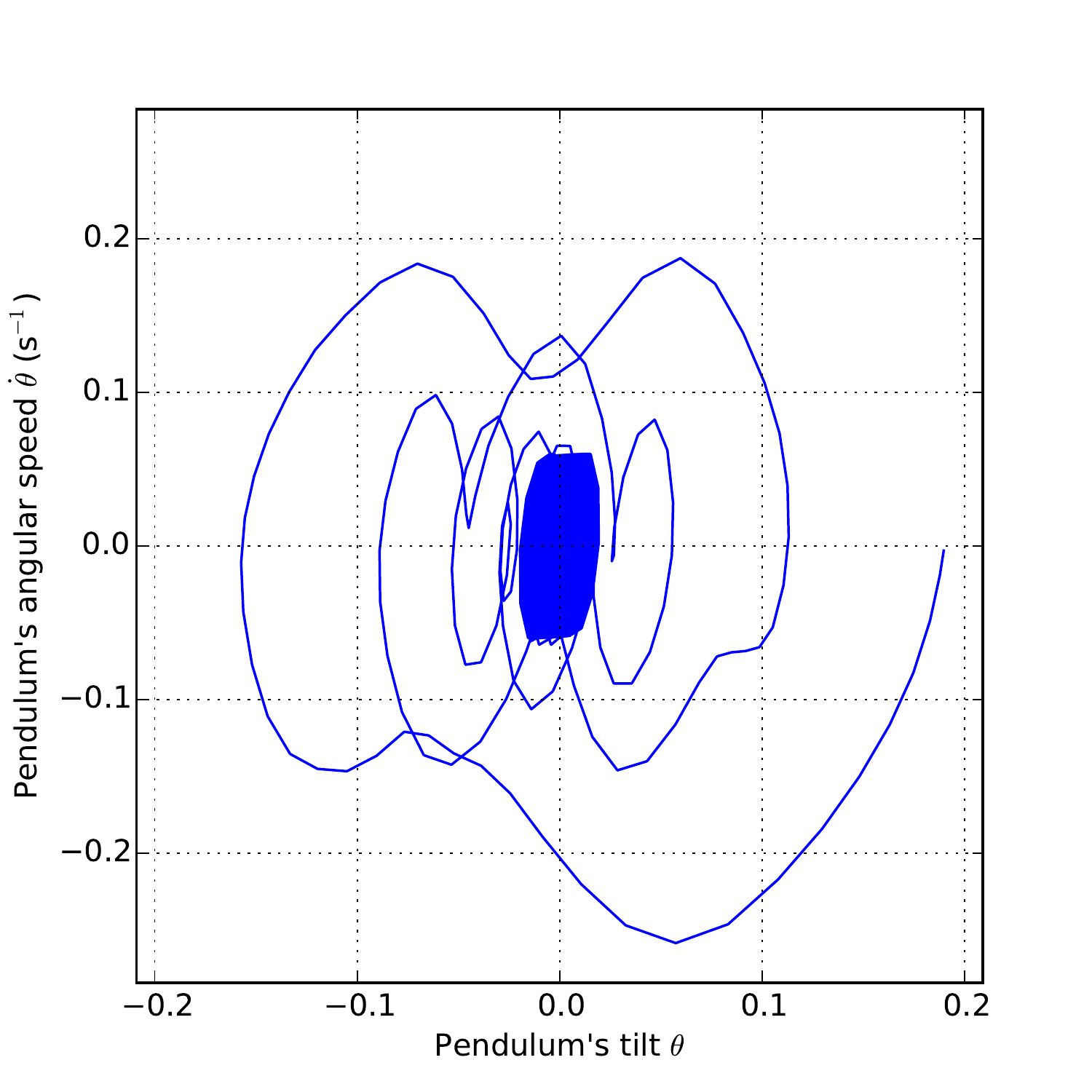}
	\caption{Inverted pendulum with derivative-free feedback ($x,\theta$), recurrent MLP
		with 20 hidden units and telescopic optimization on 16 bits per weight. Top: evolution of the positional variables in time;
		middle: network output (force applied to the cart); bottom: phase-space diagrams. Only the initial $100\text{s}$
		of simulated time are shown for clarity.}
	\label{fig:inverted nospeed recur}
\end{figure}
Fig.~\ref{fig:inverted nospeed recur} shows the behavior of the system when controlled by a recurrent network with 10 hidden units
trained with the
same parameters as before. The control variable
output by the neural net is quite jittery, possibly due to time discretization, but the resulting system can keep track of its
current status and maintain the pendulum close to the upright position, while keeping the cart in proximity to the central position.
The cart keeps its equilibrium throughout the $1000$-second simulation, but only the initial $100\text{s}$ are shown for clarity.
The limit cycles tend to drift back and forth, therefore they don't clearly appear in the two bottom phase-space diagrams.
Although the dynamics look quite different, the overall error is kept low by the lower variability of the pendulum tilt:
$\text{Err}=2.28\cdot10^{-2}$.

\subsection{Convergence to suboptimal attractors}

\begin{figure}[tb]
	\includegraphics[width=\hsize]{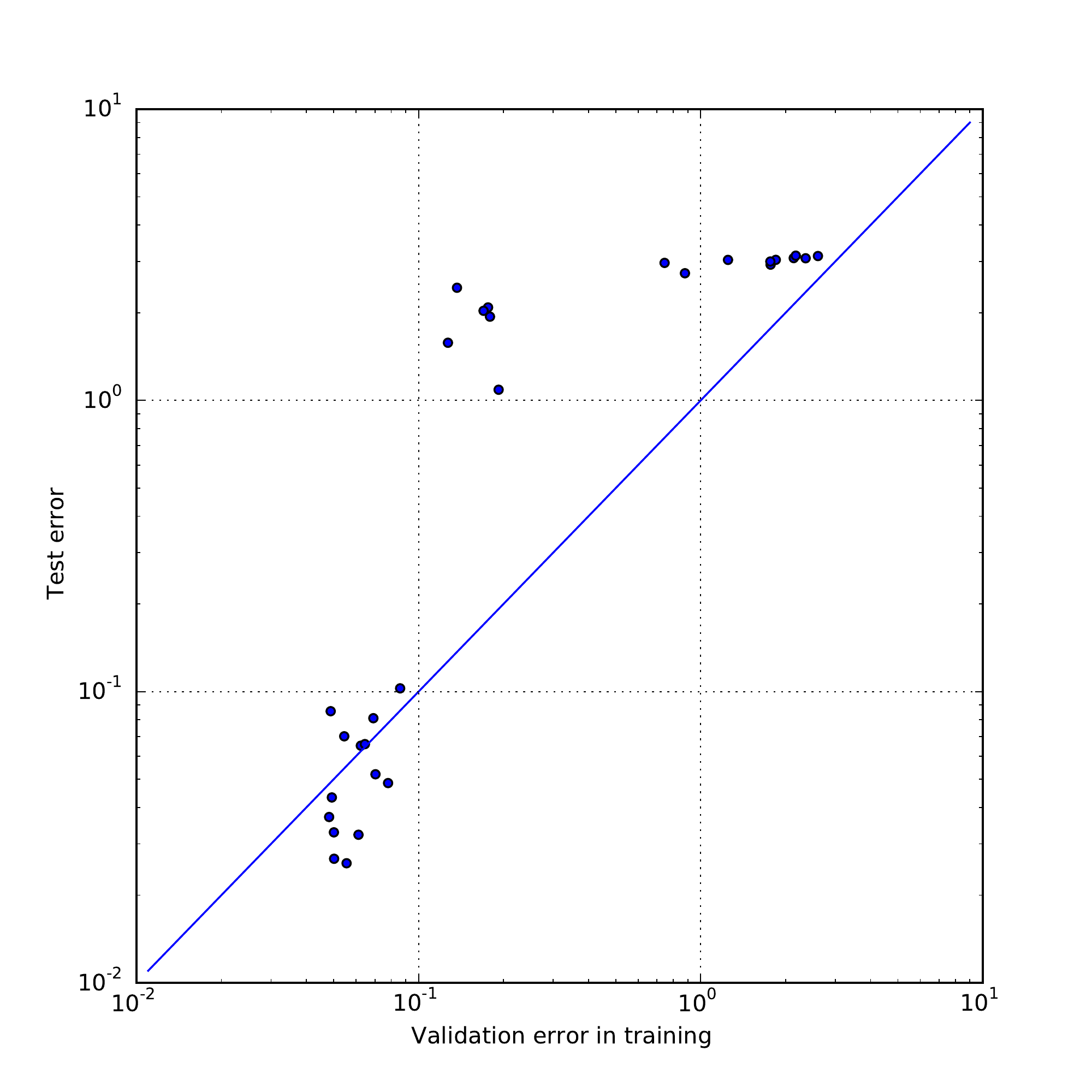}
	\caption{Distribution of validation and test results for a derivative-free instance
	of the inverted pendulum problem. Test instances
	have a longer simulated time, therefore tend to have extreme outcomes.}
	\label{fig:inverted nospeed cloud}
\end{figure}
While training recurrent networks on the derivative-free version, the existence of suboptimal attractors becomes apparent.
We performed 30 independent training sessions on
networks with 10~hidden units and 16~bits per weight. The telescopic method was used, starting from 2 bits,
and increasing the number of bits whenever the estimated ratio of improving moves falls below $\phi=10\%$,
estimated by means of the technique described in Sec.~\ref{sec:threshold} with a mobile average decay factor $\eta=.95$.
Fig.~\ref{fig:inverted nospeed cloud} shows a scatterplot of the best validation error obtained during training versus
the test error computed with a fresh set of examples and a longer simulated time.
While about half of the tests is concentrated in the low-error area ($Err<.1$), the other half converges to configurations with
much higher errors, which are amplified by the longer testing times.

Observe that test errors tend to be much more polarized (almost no values in $[0.1,1]$) than the corresponding validation
errors. This is due to the fact that test error are executed for a longer simulated timespan ($1000\text s$ vs. $100\text s$),
while training and validation instances are kept shorter in order to speed up the training phase.

\begin{table}[tb]
	\caption{Convergence to suboptimal attractors in the inverted pendulum derivative-free case (30 samples)}
	\label{tab:convergence}
	\centering
	\begin{tabular}{r|rrr|r}
		\multicolumn{1}{c|}{Hidden} & \multicolumn{3}{c|}{Test error} & \multicolumn{1}{c}{Median time} \\
		\multicolumn{1}{c|}{units} & \multicolumn{1}{c}{Minimum} & \multicolumn{1}{c}{1st quartile} & \multicolumn{1}{c|}{\# less than $.1$} & \multicolumn{1}{c}{(seconds)} \\
		\hline
		5 & 1.205 & 1.205 & 0 / 30 & 42 \\
		10 & 0.023& 0.130 & 7 / 30 & 97 \\
		20 & 0.037 & 0.052 & 13 / 30 & 222 \\
		40 & 0.028 & 0.382 & 4 / 30 & 568 \\
	\end{tabular}
\end{table}
The non-negligible ratio of suboptimal attractors can be avoided by simply repeating the optimization until the desired
error level is attained.
As shown in Tab.~\ref{tab:convergence}, in fact, apart from the smaller 5-unit network, all network sizes were able
to attain low errors in at least some of the 30 training sessions. In the 20-unit network, 13 runs out of 30 ended with a low
error: the probability of finding a good result after $N$ restarts is thus $1-(17/30)^N$, and the expected time
to obtain a low error configuration is $222\text{s}\cdot(30/13)\approx512\text{s}$.


\section{Additional tests on feed-forward networks}
\label{sec:feedforward}

\begin{table}[tbp]
    \caption{Specification of the 4 benchmark sets used in comparisons}
    \label{tab:benchmarks}
    \centering
    \begin{tabular}{lrrrr}
        Dataset name & Training samples & Test samples & Attributes & Outputs\\
        \hline
          \texttt{house16H} & 15948 & 6836 & $16$ & 1 \\
          \texttt{yeast} & 1038 & 446 & $8$  & 10 (\texttt{classif.})\\
          \texttt{abalone} & 2924 & 1253 & $7 + 1$ & 1\\
     \end{tabular}
\end{table}

In the following we collect experimental results on the widely used benchmark
real-life regression and classification problems listed in Table~\ref{tab:benchmarks}. The datasets have
been chosen among the larger ones presented in~\cite{huang2006can}.

The \texttt{house16H} dataset contains housing information that relates demographics and housing market state of a region with the median price
of houses.

The \texttt{yeast} dataset consists of a classification problem.
The task is to determine the localization site of protein in gram-negative bacteria and eukaryotic cells \cite{nakai1991expert,nakai1992knowledge}.
Output is coded with a unary representation (1 for the correct class, 0 for the wrong classes).
In addition to the traditional RMSE error, also the cross-entropy (CE) error is considered
in this classification case.

The \texttt{abalone} dataset contains data on specimens of the Abalone mollusk; 7~columns report physical measures, one column
is nominal (male/female/infant), and for our purposes it has been transformed into three $\pm1$-valued
inputs, so that a total of 10~inputs were used. The output column is the age of the specimen (number of accretion rings in the shell).

All dataset inputs have been normalized by an affine mapping of the training set inputs onto the $[-1,1]$ range and
by an affine mapping of the outputs onto the $[0,1]$ range.
The affine normalization coefficients obtained on the training set have also been applied to the test set values.
70\% of the examples are used for training, 30\% for validation.
Given that we did not repeat the experiments to determine free parameters, a separate test set was not deemed necessary (the terms
validation and test are used with the same meaning in the following description).

\subsection{House benchmark}

	

%
%

\begin{figure}[ptb]
	\includegraphics[width=\hsize]{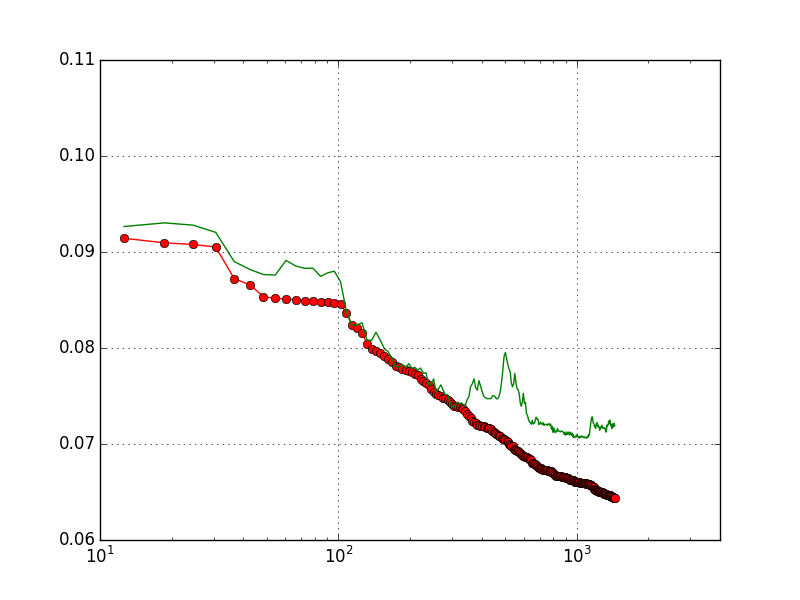}
	\includegraphics[width=\hsize]{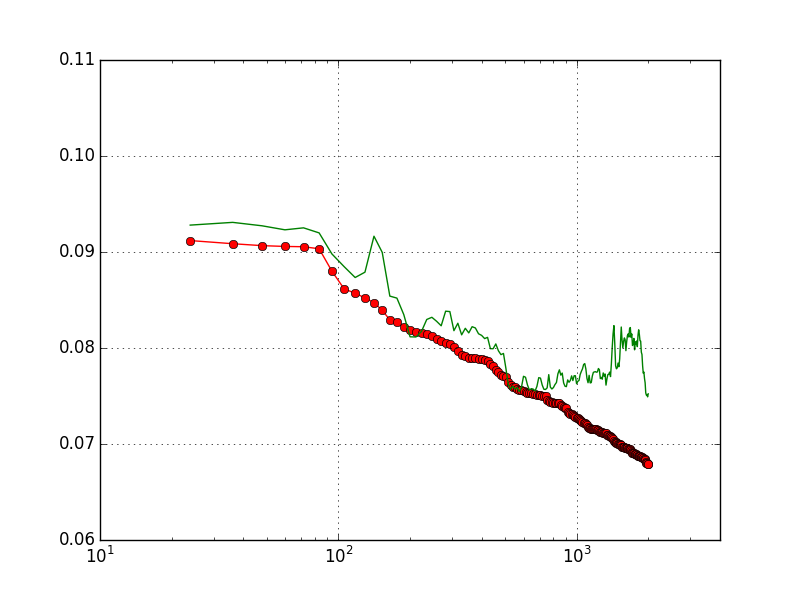}
	\caption{OSS for house: training and validation data as a function of CPU time. 100 hidden units (above), 200 hidden (below).
Average RMS error on training set (circles) and on test set (simple line).}
	\label{fig:hoss}
\end{figure}

Before analyzing the distribution of results over multiple runs, it is of interest
to observe two individual runs of OSS and BLM for 100 and 200 hidden units.
The transfer functions are 	symmetric sigmoid (hidden units) and linear (output units).
Weights are initialized randomly in the range (-0.01,0.01).
For BLM, the number of bits is 12, and the weight range 8.0.
BLM regularization weight is 1.0.
The runs last 4000 seconds.

In Fig.\ref{fig:hoss} one observes a qualitative difference: While the error on the training
set always decreases (as expected), the generalization error on the validation set
has a noisy and irregular behavior for OSS, with clear signs of overtraining at the end of the run.
The evolution of the validation error for BLM is much smoother and no sign
of over-training is present. We conjecture that this behavior is caused by
the more stochastic (less ``greedy'') choice of each weight change in BLM with
respect to OSS. The best validation results are generally better for BLM
(with improvements of around 1-3\%).

%

\begin{figure}[ptb]
	\includegraphics[width=\hsize]{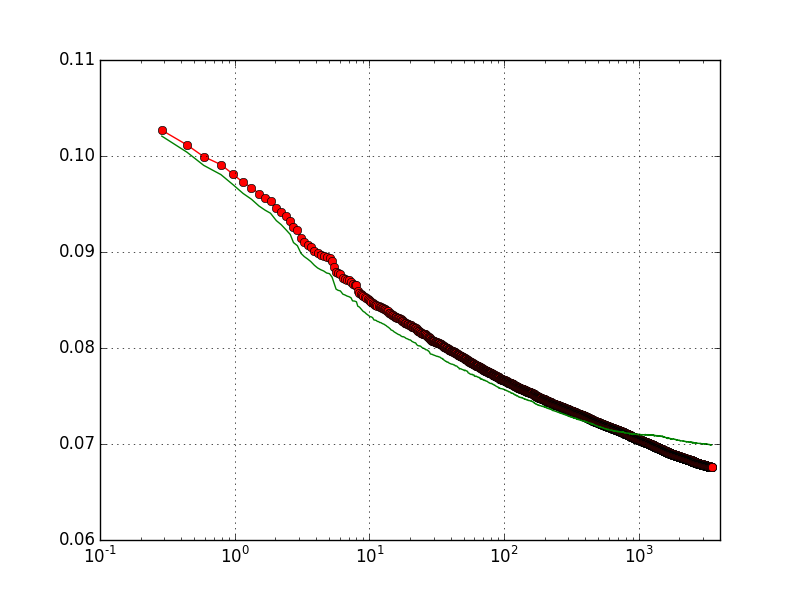}
	\includegraphics[width=\hsize]{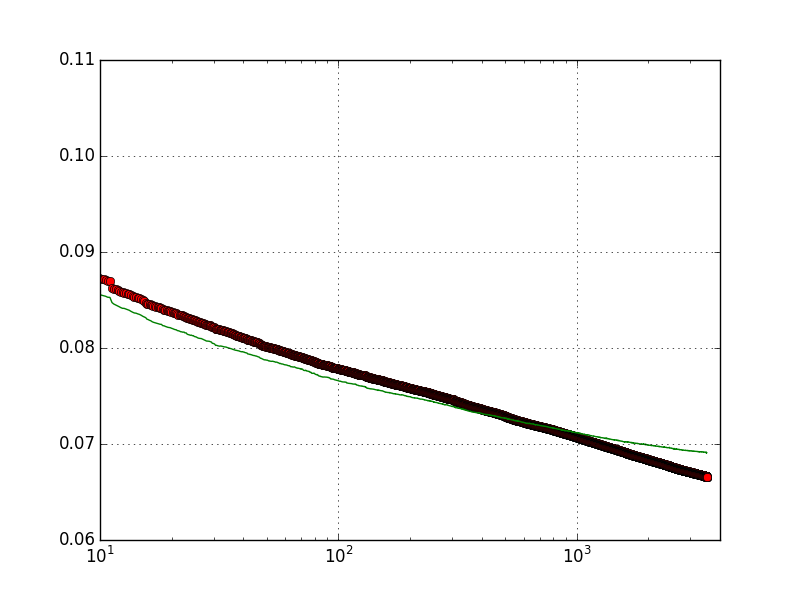}

	\caption{BLM (non-telescopic) for house: training and validation data. 100 hidden (above) 200 hidden (below).
Average RMS error on training set (circles) and on test set (simple line).}
	\label{fig:hsmooth}
\end{figure}	

	



\begin{figure}[ptb]
	\includegraphics[width=\hsize]{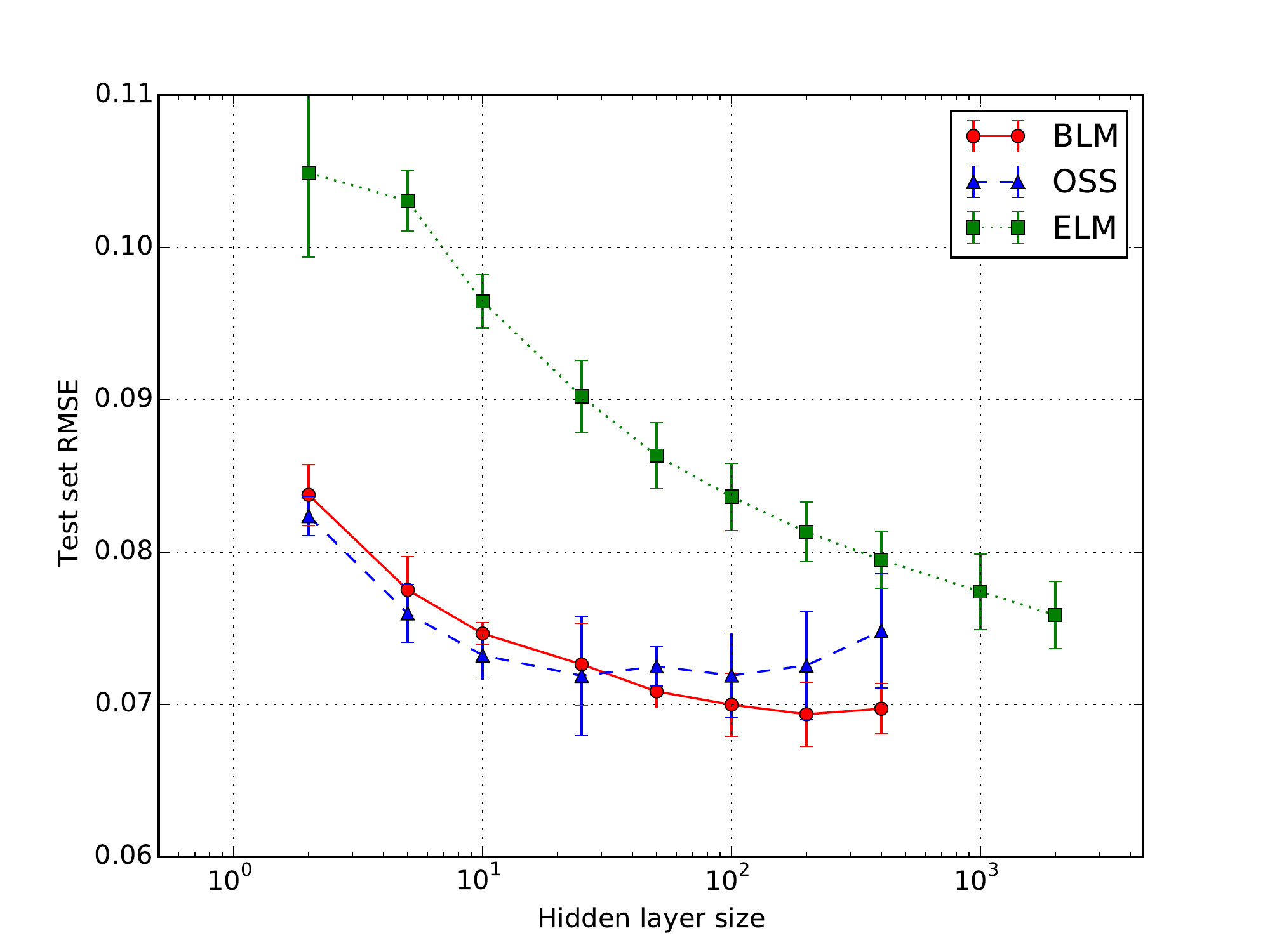}

	\caption{House benchmark. Validation results as a function of hidden layer size.}
	\label{fig:house-err}
\end{figure}


The results of five runs for each configuration, with different numbers of hidden
neurons, are plotted in Fig. \ref{fig:house-err} with error bars (estimated error
on the average).
The test results of BLM are better than those of OSS, in particular for
large networks, a result which confirms less over-training for BLM.
For 400 hidden units, results are 7\% better for BLM.
ELM needs a very large number of hidden units to obtain competitive results,
but the error for 4000 hidden nodes (0.0758) is still 10\% worse than the
best BLM results.

\begin{figure}[ptb]
	\includegraphics[width=\hsize]{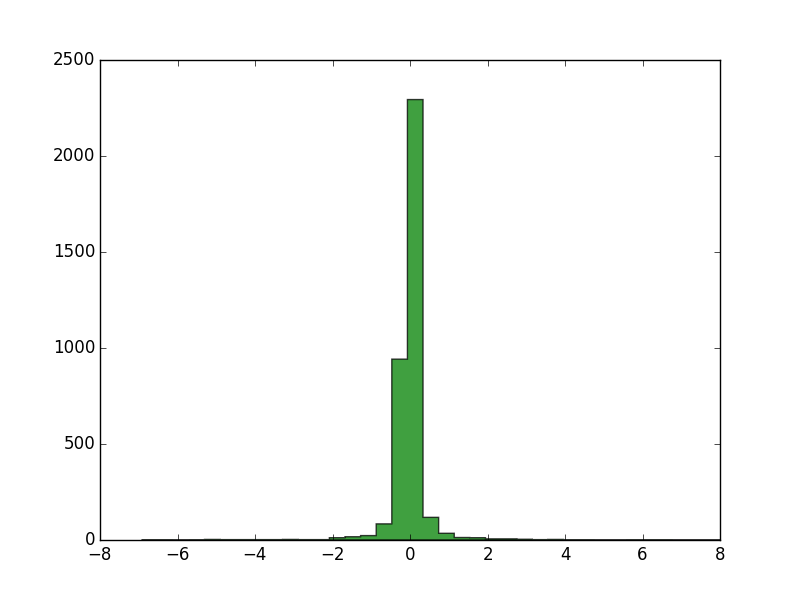}
	\includegraphics[width=\hsize]{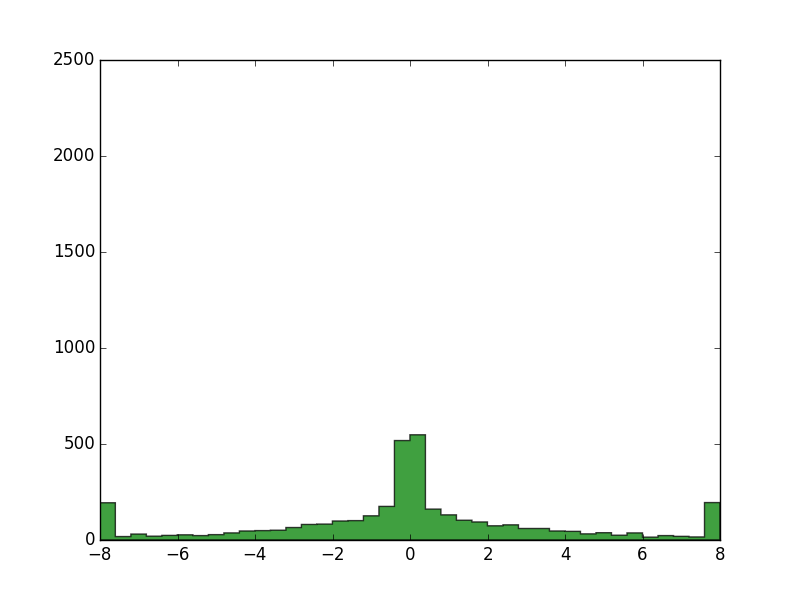}

	\caption{House benchmark. Distribution of weight values (200 hidden units).
	OSS (above) BLM (below).}
	\label{fig:wei}
\end{figure}	

In order to understand if the difference in validation is related to the final
size of weights, the final distribution
of weights is shown in Fig. \ref{fig:wei}.
The histogram shows that the better generalization for BLM cannot be explained in a simple manner by a smaller average
weight magnitude, which actually tends to be larger for BLM.
There is evidence that the search trajectory in weight space of BLM explores parts
which are not explored by the more ``greedy'' OSS based on gradient descent.

\subsection{Yeast benchmark}

%
%
%
%


The yeast benchmark task is a classification problem. With our unary output encoding (1 for the correct
class, 0 otherwise), at least two error measures are of interest, the RMS error and the
cross-entropy (CE) error. The second better reflects the classification nature of the task,
but the first is also reported for uniformity with the other benchmarks considered in this paper.

\begin{figure}[ptb]
	\includegraphics[width=\hsize]{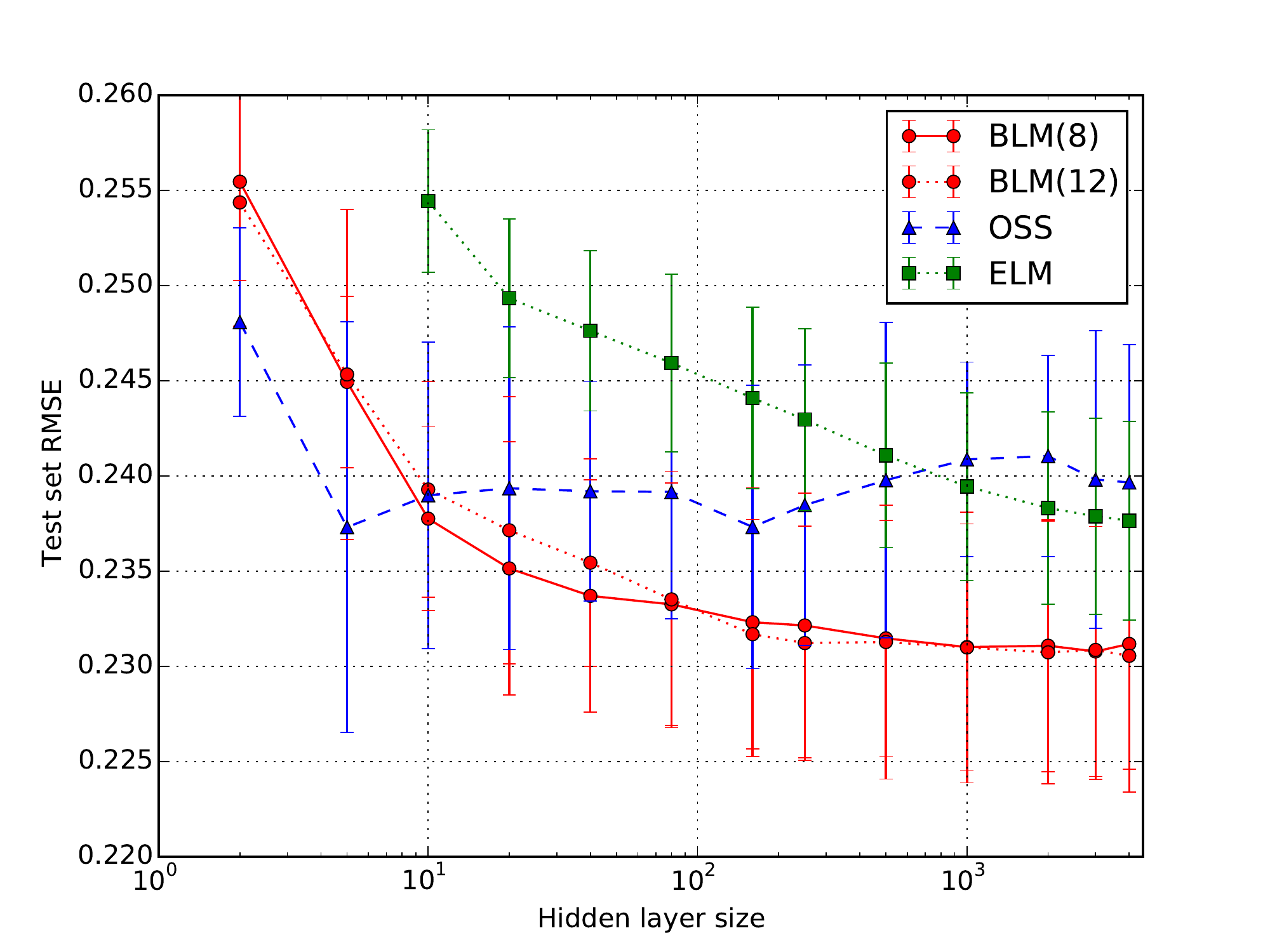}
	\caption{Yeast benchmark (RMS error). Validation results as a function of hidden layer size.
	BLM with maximum weight range 8.0 (BLM(8)) and 12.0 (BLM(12)), ELM, and OSS.}
	\label{fig:yeast-err}
\end{figure}

The parameters of the runs are:
symmetric sigmoid (hidden layer), standard 0-1 sigmoid (output layer).
Initial weigh range 0.001,
two values for the weight range in BLM (8 and 12).
The runs last 500 seconds.
The RMSE validation results (Fig. \ref{fig:yeast-err}) as a function of the number
of hidden nodes show superior results by BLM. OSS obtains close results for
small numbers of hidden units but shows a larger standard deviation.
ELM needs a very large hidden layer to reach result of interest.
For a quantitative comparison, the best average validation results are
of for 0.231 BLM, of 0.237  for OSS, of 0.238 for ELM.

The two plots for the different weight range (8 and 12) for BLM show
that the detailed choice of this parameter is not critical
(a value between 6 and 12 usually corresponds to a plateau
of best results).

%
%
%

%

\begin{figure}[ptb]
	\includegraphics[width=\hsize]{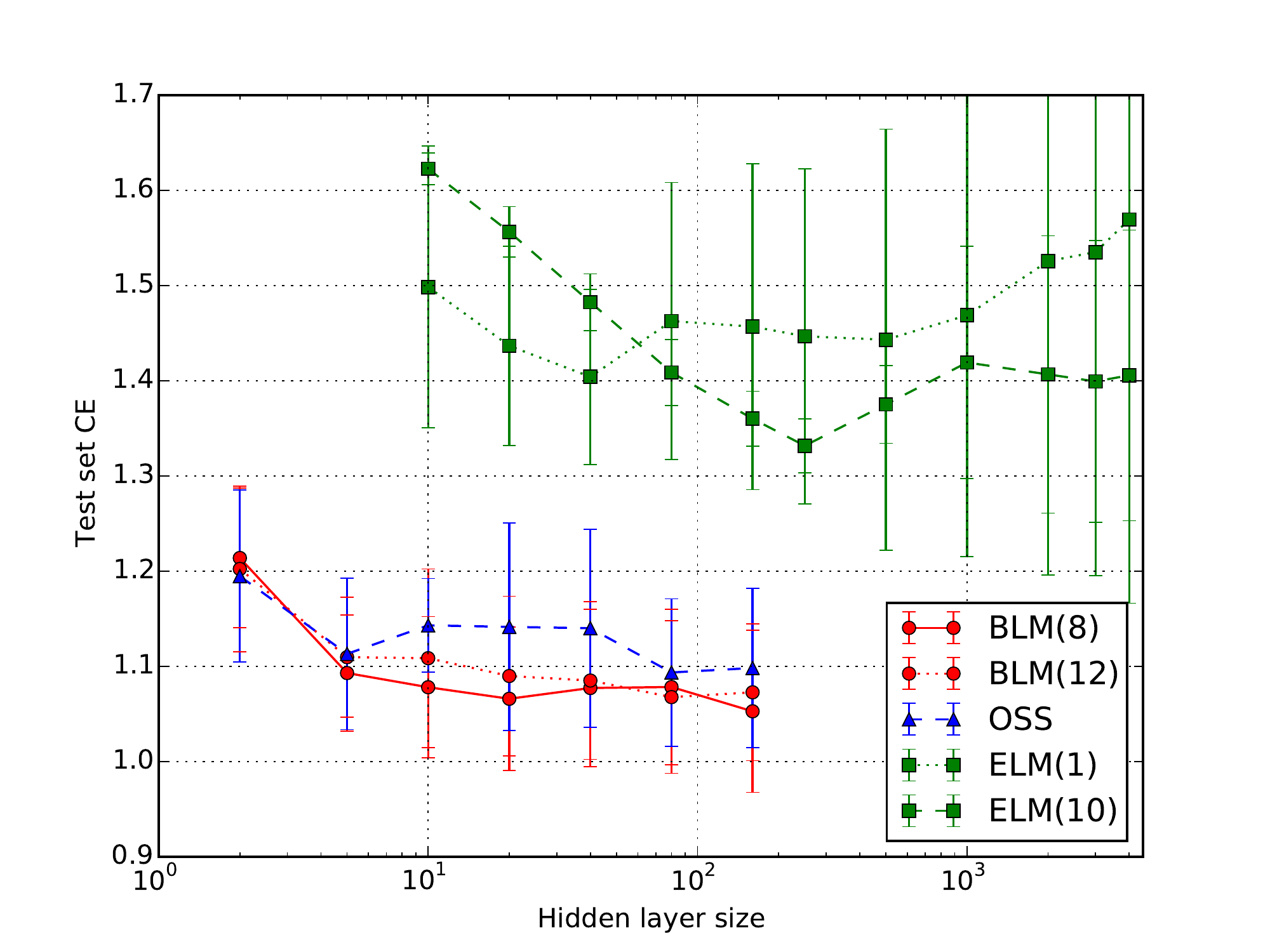}
	\caption{Yeast benchmark (CE error). Validation results as a function of hidden layer size.
	BLM with maximum weight range 8.0 (BLM(8)) and 12.0 (BLM(12)), ELM with regularization parameter 1 (ELM(1),
	with regularization parameter 10 (ELM(10)), and OSS. }
	\label{fig:yeast-ce-err}
\end{figure}

The results for the cross-entropy error (Fig. \ref{fig:yeast-ce-err}) show
similar validation results for OSS and BLM, and inferior results for ELM.
In an attempt to improve BLM results, different values of the regularization parameter
(from 0.1 to 100) have been tested. Two plots for values 1 and 10 are shown.
The parameter does influence the results but they remain in any case far from those
of BLM and OSS.

\subsection{Abalone benchmark}

\begin{figure}[ptb]
	\includegraphics[width=\hsize]{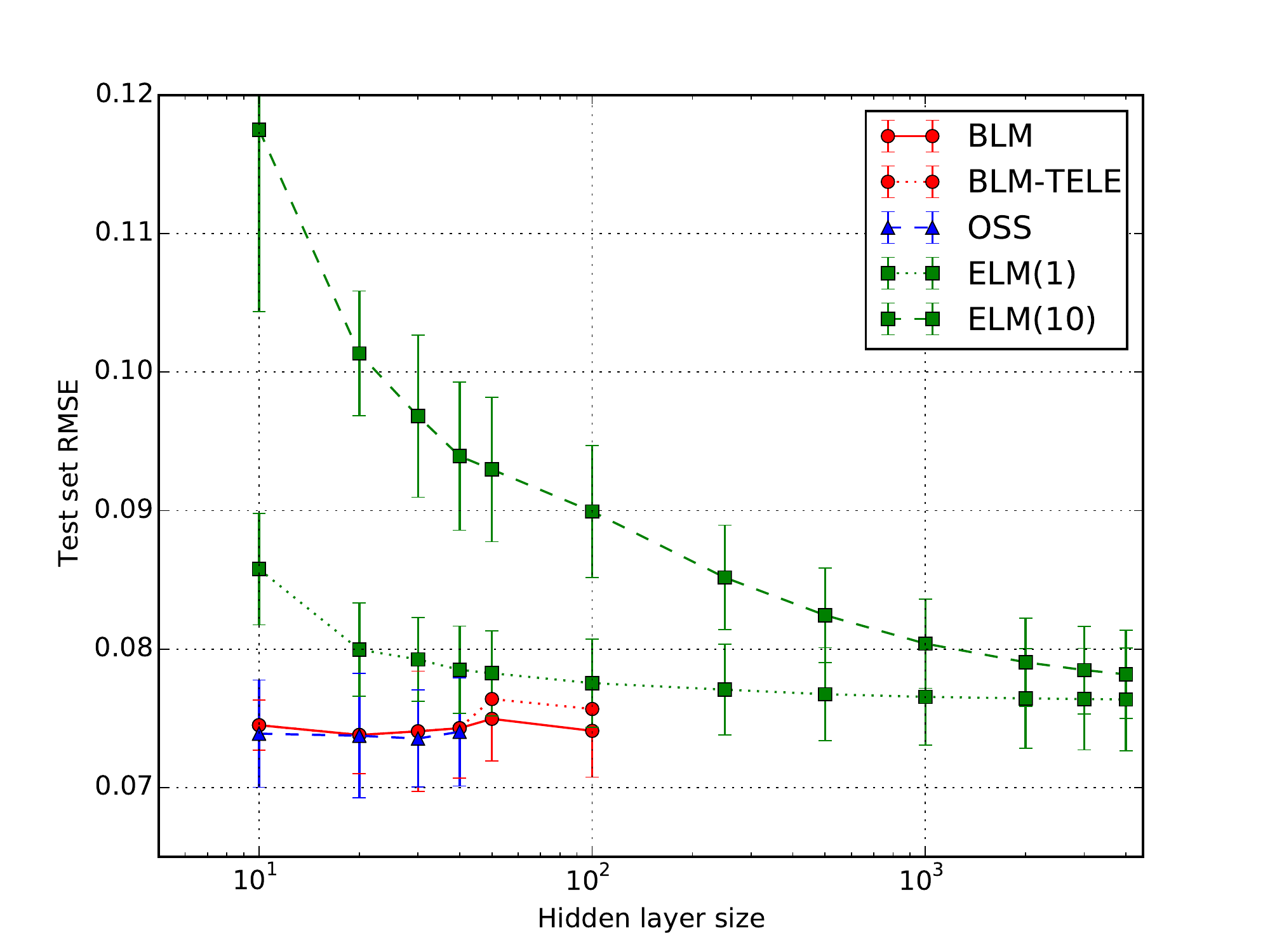}
	\caption{Abalone benchmark (RMS error). Validation results as a function of hidden layer size.
	BLM with maximum weight range 8, Telescopic BLM, ELM with regularization parameter 1 (ELM(1),
	with regularization parameter 10 (ELM(10)), and OSS. }
	\label{fig:abalone-err}
\end{figure}

The parameters of the runs are:
symmetric sigmoid (hidden layer), linear (output layer).
Initial weigh range 0.001,
two values for the regularization parameter of BLM (1 and 10).
For BLM, both the telescopic and the normal version are compared.
The runs last 500 seconds.

The average RMS error on validation for different numbers of neurons in the hidden layer
is shown in Fig. \ref{fig:abalone-err}.
One can observe that OSS and BLM results are similar. Results of
normal or telescopic BLM are almost indistinguishable.
As usual, ELM needs a fat hidden layer to reach competitive results,
but results with 4000 hidden units are still larger than those of OSS/BLM.

\section{Conclusions}
\label{sec:conclusions}

The objective of this paper was not that of proposing
yet another technique but that of revisiting
basic stochastic local search to assess the baseline
performance of algorithmic building blocks
in order to motivate more complex methods.
Better said, the algorithm considered from the point of view of
discrete dynamical systems in the weight space is simple; complexity
is delegated to the implementation with the design of efficient
methods and data structures to enormously reduce CPU times.

The results were counterintuitive. In the considered benchmark cases
our Telescopic BLM algorithm not only reproduced results obtained by more complex
methods, but actually surpassed them by a statistically significant gap
in some cases.
Also the performance produced by extreme learning was improved,
in some cases with much smaller networks (with a smaller
number of hidden units). To be fair, let's note that ELM is extremely fast
and better than simple back-propagation and therefore
should not be discounted as a promising technique.

The experimental results indicate that a simple method like
BLM based on stochastic local search and an adaptive
setting of the number of bits per weight
is fully competitive with more complex approaches
based on derivatives or based only on function evaluations.

The speedup obtained by supporting data structures, incremental
evaluations, a small and adaptive number of bits per weight,
and stochastic first-improvement neighborhood explorations
is of about two orders of magnitude for the benchmark problems
considered, and it increases with the network dimension.
The CPU training time is still much larger
than that obtained by ELM via pseudo-inverse, but still
acceptable if the application area does not require very
fast online training.

We feel that the interesting results obtained
will motivate a wider application of BLM for
more complex networks and machine learning schemes,
and we are continuing the investigation in this direction.
We encourage researchers to consider special-purpose hardware
implementations, or realizations on GPU accelerators.
Ensembling is another promising avenue
\cite{chen2007ensembling}, as well as the theoretical analysis
of the dynamics in weight space produced by Telescopic BLM.

\bibliographystyle{IEEEtran}
\bibliography{biblio}

\end{document}